\documentclass[conference]{IEEEtran}
\usepackage{times}

\usepackage[numbers]{natbib}
\usepackage{multicol}
\usepackage[bookmarks=true]{hyperref}
\usepackage{graphicx}

\usepackage{amsmath}       
\usepackage{amssymb}       
\usepackage{amsfonts}      
\usepackage{mathtools}     
\usepackage{bm}            
\usepackage{bbm}           
\usepackage{xfrac}
\usepackage{utfsym}
\usepackage{makecell}
\usepackage{times}         
\usepackage{dsfont}        
\usepackage{xspace}        
\usepackage{soul}
\usepackage[normalem]{ulem}
\setul{1.5pt}{1pt}

\usepackage{graphicx}      
\usepackage{wrapfig}       
\usepackage{float}         
\usepackage{stfloats}      
\usepackage{capt-of}       
\usepackage[font=footnotesize,labelfont=bf]{caption}      

\usepackage[dvipsnames,table,xcdraw]{xcolor} 

\definecolor{mydarkblue}{rgb}{0,0.08,0.45}
\definecolor{mydarkgreen}{RGB}{0, 139, 69}
\definecolor{mygreen2}{RGB}{0, 205, 0}
\definecolor{mybrown}{RGB}{139, 69, 19}
\definecolor{boxblue}{RGB}{79,173,234}
\definecolor{tablepeach}{RGB}{255, 240, 235}
\definecolor{tablepurple}{RGB}{248,235,252}
\definecolor{tableblue}{RGB}{235,241,255}

\usepackage{booktabs}      
\usepackage{multirow}      
\usepackage[flushleft]{threeparttable} 
\usepackage{arydshln}      
\usepackage{listings}
\usepackage{colortbl}
\usepackage{hyperref}                   
\usepackage[capitalise, nameinlink]{cleveref} 
\crefname{section}{Sec.}{Secs.}
\crefname{table}{Tab.}{Tabs.}

\usepackage{xurl}                       
\definecolor{citecolor}{HTML}{0a29c2} 

\hypersetup{
    colorlinks=true,
    linkcolor=citecolor,
    urlcolor=black,
    citecolor=citecolor,
}
\usepackage[linesnumbered,ruled,vlined]{algorithm2e} 

\usepackage{pifont}  
\usepackage{dsfont}  
\usepackage{lipsum}  
\usepackage{siunitx} 
\usepackage{multicol} 
\let\labelindent\relax
\usepackage{enumitem}

\newcommand{\ours}{{TAG}\xspace}
\newcommand{\ourrow}{\rowcolor{gray!7}}

\newcommand{\yes}{\large \color{OliveGreen}\checkmark}
\newcommand{\no}{\color{BrickRed} \scalebox{1}{\usym{2613}}}

\newcommand{\half}{\large \color{orange} \ooalign{\checkmark\cr\hidewidth\kern 0.2em\raisebox{0.3ex}{\scalebox{0.7}{\pmb{\textbackslash}}}\hidewidth}}

\begin{document}

\title{Feel Robot Feels: Tactile Feedback Array Glove\\ for Dexterous Manipulation}
\author{\authorblockN{Feiyu Jia\textsuperscript{2,1,*} \quad Xiaojie Niu\textsuperscript{1,*}
\quad Sizhe Yang\textsuperscript{1,3,*}
\quad Qingwei Ben\textsuperscript{1,3}  \quad Tao Huang\textsuperscript{4,1}  \\ \quad Feng Zhao\textsuperscript{2,$\dagger$} \quad Jingbo Wang\textsuperscript{1,$\dagger$} \quad Jiangmiao Pang\textsuperscript{1,$\dagger$}
}
\authorblockA{
\textsuperscript{1} Shanghai AI Laboratory \quad 
\textsuperscript{2} University of Science and Technology of China \quad \\
\textsuperscript{3} The Chinese University of Hong Kong \quad  
\textsuperscript{4} Shanghai Jiao Tong University \\
\textsuperscript{*} Equal Contributions \quad 
\textsuperscript{$\dagger$} Corresponding Authors \\
}}




%

\twocolumn[{%
\renewcommand\twocolumn[1][]{#1}%
\maketitle
\vspace{-0.4cm}
\begin{center}
    \centering
    \captionsetup{type=figure}
     \includegraphics[width=1.0\textwidth]{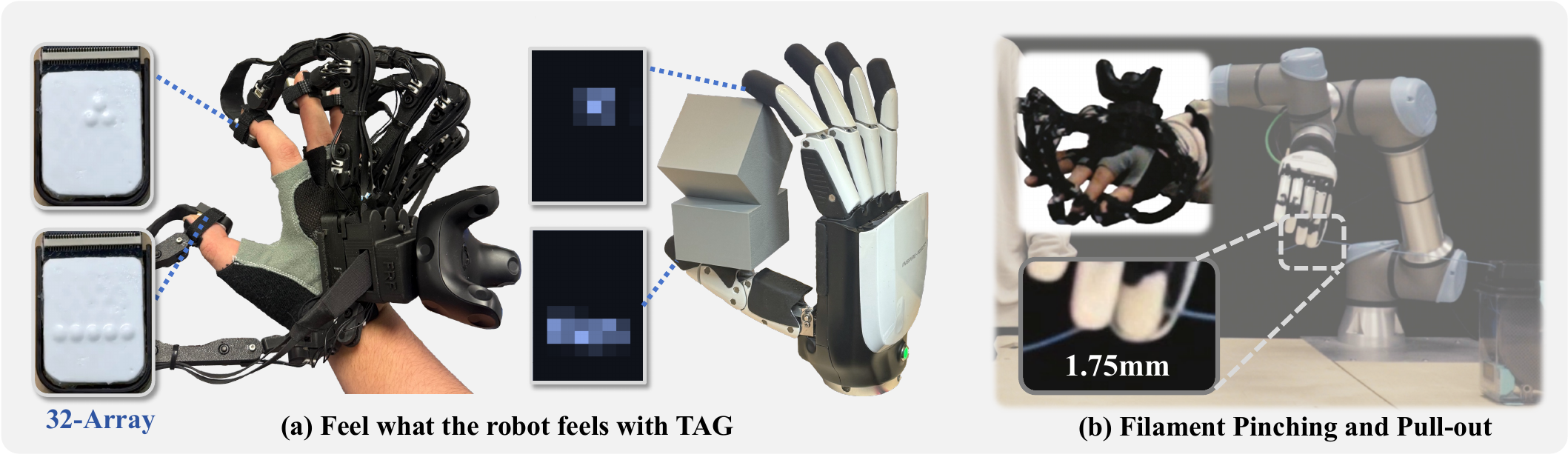}
     \vspace{-0.17in}
    \caption{(a) \textbf{TAG} is a \textbf{\ul{T}}actile Feedback \textbf{\ul{A}}rray \textbf{\ul{G}}love for dexterous manipulation and teleoperation. It achieves precise 21-DoF joint tracking and integrates a 32-point  feedback array on each fingertip, bridging the gap between robot-side sensing and human-side perception. (b) Sequential filament pinching tack. } 
    \label{fig:teaser}
\end{center}
\vspace{0.04in}
}]

\begin{abstract}

   Teleoperation is a key approach for collecting high-quality, physically consistent demonstrations for robotic manipulation. However, teleoperation for dexterous manipulation remains constrained by: (i) inaccurate hand–robot motion mapping, which limits teleoperated dexterity, and (ii) limited tactile feedback that forces vision-dominated interaction and hinders perception of contact geometry and force variation.
   To address these challenges, we present TAG, a low-cost glove system that integrates precise hand motion capture with high-resolution tactile feedback, enabling effective tactile-in-the-loop dexterous teleoperation.  For motion capture, TAG employs a non-contact magnetic sensing design that provides drift-free, electromagnetically robust 21-DoF joint tracking with joint angle estimation errors below 1°. Meanwhile, to restore tactile sensation, TAG equips each finger with a 32-actuator tactile array within a compact 2 cm² module, allowing operators to directly feel physical interactions at the robot end-effector through spatial activation patterns.
   Through real-world teleoperation experiments and user studies, we show that TAG enables reliable real-time perception of contact geometry and dynamic force, improves success rates in contact-rich teleoperation tasks, and increases the reliability of demonstration data collection for learning-based manipulation.
   Videos and code are available on our \href{https://trap-1.github.io/TAG.github.io/}{\textcolor{citecolor}{project page}}.
   
\end{abstract}

\IEEEpeerreviewmaketitle

\section{Introduction}
\label{sec:intro}
Dexterous manipulation is a core challenge in robotics and is increasingly addressed via learning from real-world demonstrations~\cite{wen2025gr,pi0,dp,dp3,act}. The success of these data-driven methods depends critically on the fidelity and coverage of the collected data~\cite{generalist2025gen0}. This dependence has driven the development of recent teleoperation systems~\cite{yang2025ace,cheng2024open,ben2025homie,zhang2025do}, which offer practical and scalable interfaces for data collection.


Vision- and Virtual Reality (VR)-based teleoperation systems are widely used but are limited by visual blockage, pose estimation errors, and viewpoint misalignment~\cite{DexPilot,cheng2024open}. Glove-based approaches bypass these vision-dependent constraints through direct motion sensing, yet still face challenges in signal reliability: IMU-based systems suffer from drift over time, while contact-based mechanical sensors~\cite{zhang2025do,flexsensor} exhibit nonlinearity that increases with fatigue. Non-contact magnetic sensing~\cite{ben2025homie} alleviates fatigue-related issues, but is typically limited to single-axis measurements, preventing precise angular estimation. Although commercial solutions~\cite{manus2026} offer higher precision, they are both expensive and sensitive to electromagnetic interference. Consequently, reliable demonstration collection calls for a motion capture glove that provides both robust sensing and precise angular accuracy.


Beyond motion tracking, another critical challenge in teleoperation is the absence of haptic feedback. Without tactile cues, operators face a significant sensory gap, relying passively on vision and struggling to perceive object geometry or contact forces. While traditional pneumatic and mechanical systems offer strong feedback, their bulkiness severely restricts dexterity. Conversely, vibrotactile motors~\cite{ding2024bunny, zhang2025do} are lightweight but fail to render complex surface geometries. Recently, Electro-Osmotic Pumps (EEOPs) have emerged as a promising solution, utilizing electro-hydrodynamics to generate high pressure in a silent, millimeter-scale form factor; however, current applications~\cite{FluidReality} are confined to tethered VR simulations. Therefore, realizing high-fidelity tactile feedback for real-world robotic teleoperation remains an unsolved challenge.

In this paper, we introduce \ours, a low-cost glove system that features precise hand motion capture with high-resolution tactile feedback, enabling effective dexterous tactile-in-the-loop teleoperation. By deploying a non-contact magnetic tracking framework across a strategic 21-DoF configuration, our system guarantees drift-free joint estimation, enabling highly precise and stable robot manipulation.

To bridge the sensory gap, each fingertip is equipped with a compact 2 cm² module housing a 32-actuator EEOP tactile array. We optimize channel ratios and grid symmetry to ensure high fluid flux and consistent spatial resolution, while relaxing fabrication tolerances to enable handcrafted assembly. Through this high-resolution feedback module, we translate robotic data into human-perceivable cues via two distinct mapping strategies: a spatial mapping that encodes contact geometry into 2D actuator patterns for contour perception, and a pressure mapping that translates force magnitude into active contact area expansion for intuitive force feedback.


We validate \ours's tactile-in-the-loop perceptual capabilities and its effectiveness for dexterous teleoperation through real-world experiments. Quantitative results show that TAG achieves sub-degree joint angle accuracy while remaining robust to electromagnetic interference, enabling reliable hand–robot motion retargeting. User studies further confirm that the rendered tactile feedback allows operators to consistently perceive contact geometry and dynamic force direction during manipulation. Teleoperation experiments on real robotic platforms demonstrate improved performance in contact-rich tasks that challenge vision-dominated control. Finally, imitation learning results indicate that demonstrations collected with TAG are reliable and physically consistent, making them suitable for learning-based manipulation.

The main contributions of this work are three-fold:
\begin{itemize}
    \item We develop a robust 21-DoF magnetic tracking system that achieves drift-free, sub-degree precision for reliable demonstration collection.
    \item We construct high-resolution EEOP tactile modules and integrate them into real-world robotic loops, employing a dual-mode strategy to render geometry and force.
    \item We propose \ours{} and validate it on real-world challenging contact-rich tasks, demonstrating its superior capability in fine-grained teleoperation.
\end{itemize}

\section{Related Work}
\label{sec:rel}
\subsection{Data collection for Dexterous Hands}
Large-scale, high-quality demonstrations are essential for learning autonomous dexterous manipulation policies~\cite{feng2025learning,heng2025vitacformer,lin2025learning}. One prevalent paradigm records human hand motions directly using vision-based pose estimation~\cite{sivakumar2022robotic,wang2023mimicplay}, or wearable devices like exoskeletons and gloves~\cite{xu2025dexumi,fang2025dexop,wang2024dexcap}.  
However, vision-based methods are limited by pose estimation accuracy and are highly susceptible to occlusion, while exoskeletons are often tightly coupled to specific hand kinematics. More critically, transferring human motion to robot hands requires complex kinematic mapping, which inevitably introduces an embodiment gap and action–vision mismatch. 
To ensure action–observation consistency, other works teleoperate the robot hand using vision-based pose estimation (VR~\cite{cheng2024open,lu2025mobile} or camera~\cite{DexPilot,anyteleop}) or motion-capture gloves. Vision-based methods typically rely on inaccurate hand pose estimation and provide little physical contact feedback, limiting data quality in contact-rich tasks, while gloves suffer from sensor limitations. Specifically, resistive sensors~\cite{zhang2025do, flexsensor} exhibit mechanical fatigue and non-linearity; IMUs~\cite{imutele,Harrison2024imu} drift over time; and single-axis Hall sensors~\cite{ben2025homie} lack precision. Commercial magnetic solutions~\cite{manus2026} are drift-free but costly and sensitive to electromagnetic interference.
\ours addresses these issues with robust, high-precision magnetic encoding and direct tactile feedback, enabling scalable, high-fidelity data collection.

\subsection{Tactile Feedback Gloves}
Tactile information is indispensable for dexterous manipulation~\cite{si2024difftactile}; yet, most teleoperation systems offer sparse feedback, rendering contact-rich tasks challenging despite available onboard sensing~\cite{si2022taxim,yuan2017gelsight}. To bridge this gap, prior works have introduced feedback via joint-level forces~\cite{forceglove, fang2025dexop,romero2024eyesight} or vibrotactile cues~\cite{zhang2025do,ding2024bunny}. However, these modalities suffer from inherent limitations: force feedback is predominantly kinesthetic rather than cutaneous, providing indirect cues while often restricting natural motion; vibrotactile feedback typically reduces complex contact geometries to binary signals. Similarly, electrotactile stimulation methods~\cite{VizcayElec,JiangElec} often induce unnatural sensations or suffer from instability due to varying skin impedance. While some commercial gloves achieve multi-point tactile feedback, they remain prohibitively expensive for widespread adoption. \ours embeds high-resolution electrochemical tactile modules~\cite{FluidReality} and integrates them into real-world robotic loops, employing a dual-mode strategy to render both geometry and force.

\section{Hardware Design}
\label{sec:hardware}
This section presents the hardware architecture of \ours, describing how its design principles (\cref{subsec:design_principle}) guide the kinematic structure (\cref{subsec:kinematic_design}) and the tight integration of high-precision joint tracking (\cref{subsec:joint_angle_encoders}) with high-density tactile feedback (\cref{subsec:tac-eval}) for dexterous teleoperation.

\begin{figure*}[!t]
  \centering
  \includegraphics[width=0.98\textwidth]{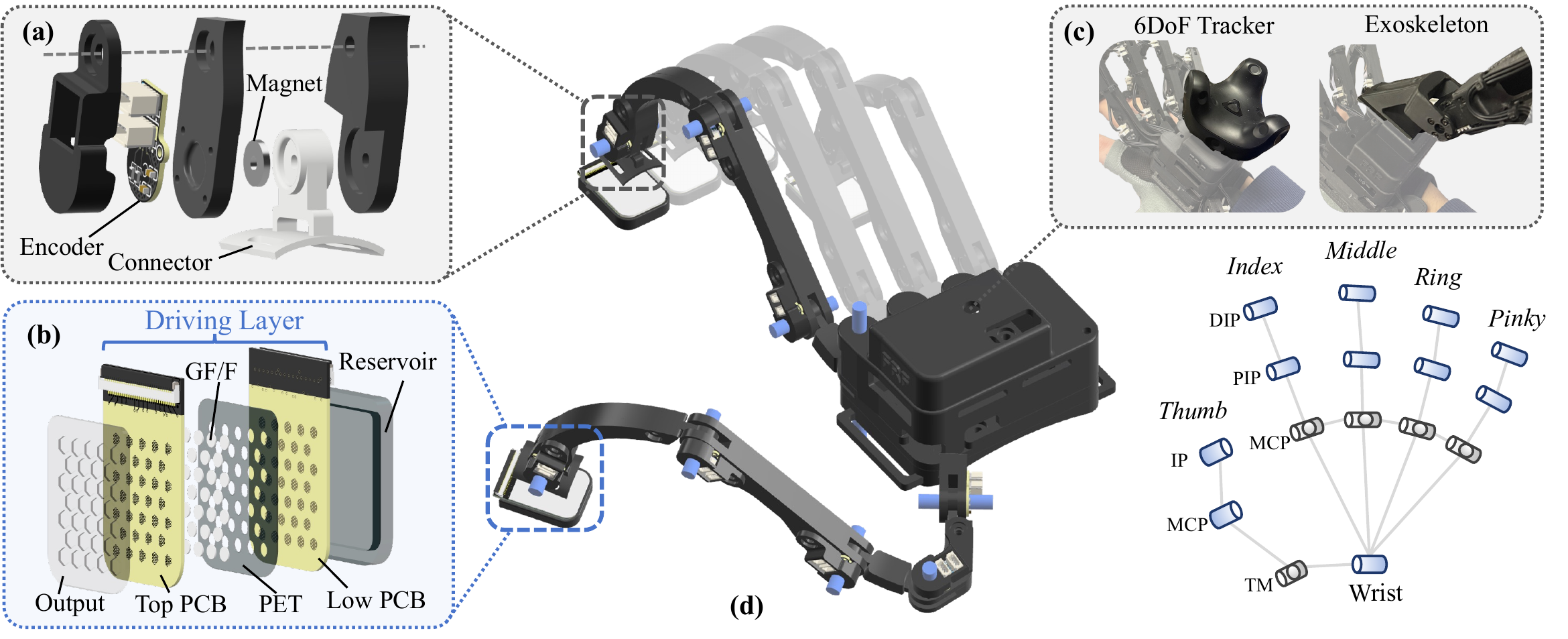}
  \caption{\textbf{Hardware architecture of \ours.} (a) Exploded view of the joint encoder, utilizing a magnetometer and ring magnet for angle sensing. (b) Exploded view of the tactile feedback module, a multi-layer EOP stack sandwiching a membrane between PCBs. (c) \ours is compatible with diverse arm tracking solutions, such as 6DoF trackers or exoskeletons. (d) Kinematic configuration distributing 21 encoders for full-hand motion capture.}
 \label{fig:design}
 \vspace{-0.15in}
\end{figure*}

\subsection{Design Principles}
\label{subsec:design_principle}
\subsubsection{\textbf{Efficient}}
\ours achieves sub-degree joint tracking accuracy ($<$$0.8^\circ$) with strong electromagnetic interference (EMI) resilience using non-contact magnetic encoders. Combined with a 32-point tactile feedback array, the system enables low-latency, high-fidelity teleoperation.

\subsubsection{\textbf{Cross-platform}}

\ours interfaces seamlessly with diverse robotic hands, tactile sensing schemes, and end-effector pose tracking methods, enabled by its high kinematic freedom and tactile sensing fidelity.


\subsubsection{\textbf{Low-cost}}
Existing commercial haptic gloves are typically expensive (over $\$5,000$) for widespread research~\cite{haptx2026, manus2026}. In contrast, \ours lowers this barrier by offering accurate full-hand motion capture and high-resolution tactile feedback at a total cost below $\$500$.

\subsubsection{\textbf{Open-source}}
To foster community-driven research, {\ours} is fully open-source. We provide comprehensive CAD models, PCB designs, and fabrication guides, ensuring that the system can be easily reproduced using standard materials.


\subsection{Kinematic Design}
\label{subsec:kinematic_design}
To achieve anthropomorphic and high-fidelity hand tracking, the kinematic structure of \ours must closely replicate the biological degrees of freedom (DoF) of a human hand. This high-fidelity capture is essential for acquiring precise demonstrations for dexterous manipulation tasks. Based on this principle, we implement a 21-DoF configuration by strategically distributing 21 magnetic encoder modules across all key anatomical joints, as illustrated in \cref{fig:design}(d).

For the four fingers (index to pinky), each is equipped with four encoders to monitor the Distal Interphalangeal (DIP), Proximal Interphalangeal (PIP), and Metacarpophalangeal (MCP) joints. Specifically, the MCP tracking is decoupled into flexion/extension and adduction/abduction to capture the complex spreading and grasping motions of the palm.

The thumb, as the most critical component for dexterous manipulation, requires an expansive range of motion to facilitate opposition, circumduction, and fine-grained pinching. Consequently, we assign 5 DoF to the thumb by deploying one encoder at the Interphalangeal (IP) joint, one at the Metacarpophalangeal (MCP) joint, and three encoders at the complex Trapeziometacarpal (TM) joint to precisely track its three-dimensional rotation. This 21-DoF design ensures a precise mapping between the operator and diverse robotic end-effectors, effectively minimizing information loss during complex teleoperation tasks.

\subsection{Joint Angle Encoders}
\label{subsec:joint_angle_encoders}

Precise joint tracking is fundamental to effective teleoperation. Unlike conventional designs that rely on potentiometers~\cite{zhang2025do} or single-axis Hall sensors~\cite{ben2025homie}, which are susceptible to mechanical wear and installation misalignment, \ours adopts a robust non-contact sensing approach utilizing 21 MLX90393 magnetometers. Selected for its high-performance Triaxis technology, this sensor resolves the magnetic flux density vector with a built-in 16-bit ADC, offering high linearity (typically $\pm 0.1\%$) and direct digital output that eliminates complex external circuitry.

As illustrated in~\cref{fig:design}(a), a miniature, diametrically magnetized Neodymium Iron Boron (NdFeB) ring magnet is positioned directly above the sensor, aligned parallel to the sensor's $XY$ plane. As the joint rotates, the sensor captures the orthogonal components of the magnetic flux density vector $\mathbf{B}$. This relationship is modeled as:
\begin{equation}
    \begin{cases}
        B_x = B_0 \cos(\theta) + O_x \\
        B_y = B_0 \sin(\theta) + O_y
    \end{cases},
\end{equation}
where $B_0$ is the magnetic flux amplitude, $\theta$ is the rotation angle, and $O_{x,y}$ represent magnetic offsets induced by environmental bias. In the \ours encoder, these offsets are calibrated by rotating the joint through its full range: 
\begin{equation}
    O_i = (B_{i,\textrm{max}} + B_{i,\textrm{min}})/2,\,\,i \in \{x, y\}.
\end{equation}

The angular position is then analytically derived using the four-quadrant arctangent function:
\begin{equation}
    \theta = \operatorname{atan2}(B_y - O_y, \, B_x - O_x).
\end{equation}
Crucially, since this calculation relies on the ratio of field components, common-mode errors, such as those induced by temperature changes, are naturally cancelled out.

Since the mechanical structure of the glove is fixed, we can accurately calculate the 3D pose of each fingertip and retarget these to various dexterous hands. Detailed implementation of the retargeting algorithms is provided in the \textcolor{citecolor}{Appendix}.

\begin{figure}[!ht]
  \centering
  \includegraphics[width=0.4\textwidth]{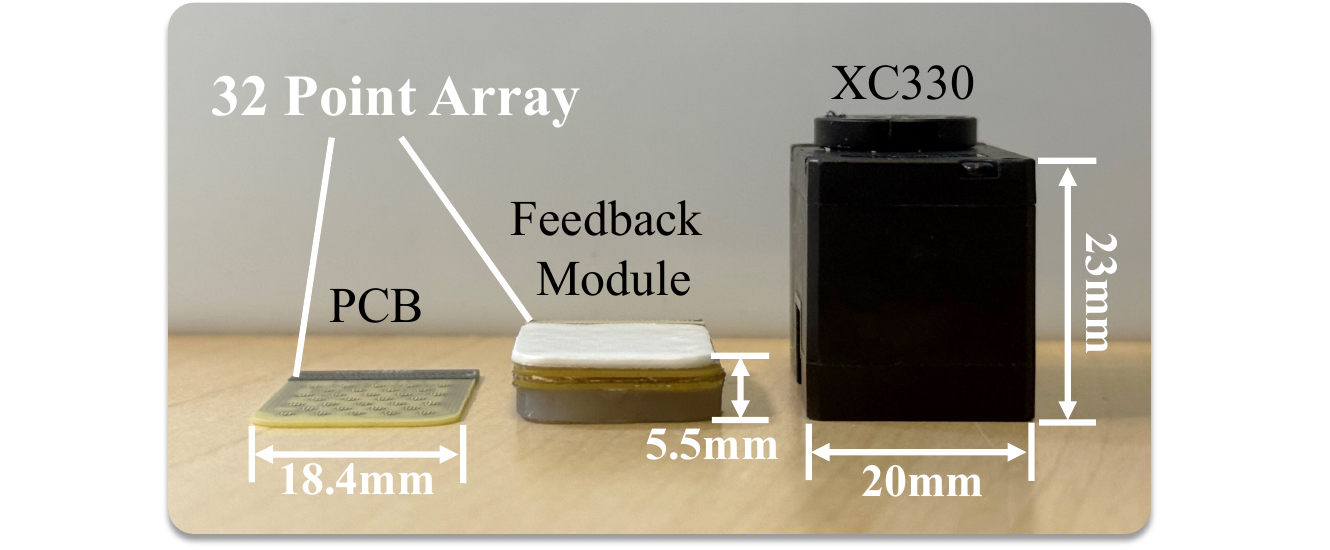}
    \caption{\textbf{Physical dimensions and form factor of the \ours feedback module.} The entire assembly measures only $29 \times 18.4 \times 5.5$ mm.}
  \label{fig:EOP_size}
    \vspace{-0.15in}
\end{figure}

\subsection{Tactile Feedback Module}
\label{sec:tactile_module}
\subsubsection{\textbf{Module Design}}
\label{subsec:tactile_design}
To deliver high-fidelity, multi-point tactile feedback within a compact $2\,\text{cm}^2$ fingertip area, we develop a tactile feedback module driven by Electro-Osmotic Pumps (EOPs). The module’s architecture is conceptually similar to prior works~\cite{FlatPanel,FluidReality,MorphingSkin} and consists of three primary layers: an output layer, a driving layer, and a reservoir layer. The output and reservoir components are cast from Ecoflex™ 00-30 using 3D-printed precision molds. The whole module is illustrated in \cref{fig:EOP_size}.

The core driving mechanism, illustrated in \cref{fig:design}(b), utilizes a glass fiber (GF/F) membrane sandwiched between dual-layer PCBs. The \ours tactile feedback module is powered by a 12V source (either a battery or external supply), which is stepped up to the required high voltage 200\,\text{V} via a boost converter. By modulating the potential difference applied to each independent channel, we enable Tri-state control for each taxel: active \textit{protrusion} (+), \textit{neutral} (0), and \textit{retraction} (-).

Building on this architecture, our implementation introduces targeted design optimizations to enhance hydraulic efficiency, spatial resolution, and manufacturability:
\begin{itemize}[leftmargin=*]
    \item \underline{Enhanced hydraulic efficiency:} We enlarge the dimensions of the \textit{vias} to $0.25/0.4\,\text{mm}$ (inner/outer). With 12 vias arranged within each $1.6\,\text{mm}$-diameter feedback taxel, this design yields a conductive-to-surface area ratio of $\sim29.3\%$.  This optimization significantly enhances hydraulic efficiency, enabling us to lower the driving voltage from 250\,\text{V} to 200\,\text{V} while maintaining high-quality fluid flux, thereby effectively reducing power consumption.
    \item \underline{Optimized spatial perception:} Prioritizing alignment with robotic sensing profiles over anatomical mimicry, we adopt a centrally symmetric layout. Unlike human fingertips which taper distally, robotic end-effectors typically feature uniform sensing arrays; our symmetric design ensures spatial mapping and consistent tactile resolution across the active area.
    \item \underline{Simplified manufacturing:} We intentionally increase the inter-node spacing to relax fabrication tolerances. This geometric adjustment eliminates the dependency on high-precision industrial tools like laser cutters, ensuring the entire assembly can be handcrafted.
\end{itemize}
Comprehensive details regarding the fabrication process and electronic control circuitry are provided in the \textcolor{citecolor}{Appendix}.

\subsubsection{\textbf{Tactile Feedback Mapping}}
\label{subsec:tactile_dmapping}
To bridge the gap between robot-side sensing and human-side perception, we develop two distinct mapping strategies for the 32-point tactile array, tailored to different task requirements.

\textbf{Shape Mapping Mode:} Intended for tasks demanding high geometric discrimination, this mode preserves the spatial topology of the contact. Raw tactile data from the end-effector is processed via bilinear interpolation and normalized to align with the staggered layout of the 32-point tactile feedback array. The mapping function maintains the spatial distribution of the input—distinguishing, for instance, between the diffuse patch of a broad contact and the narrow profile of a sharp edge. As illustrated in \cref{fig:sensor_eva}, the activation state of each array cell $k \in \{0, \dots, 31\}$ is driven by the normalized intensity at its corresponding mapped coordinate, ensuring a direct spatial correspondence for perceiving object shapes.
\begin{figure}[!ht]
  \centering
  \includegraphics[width=0.45\textwidth]{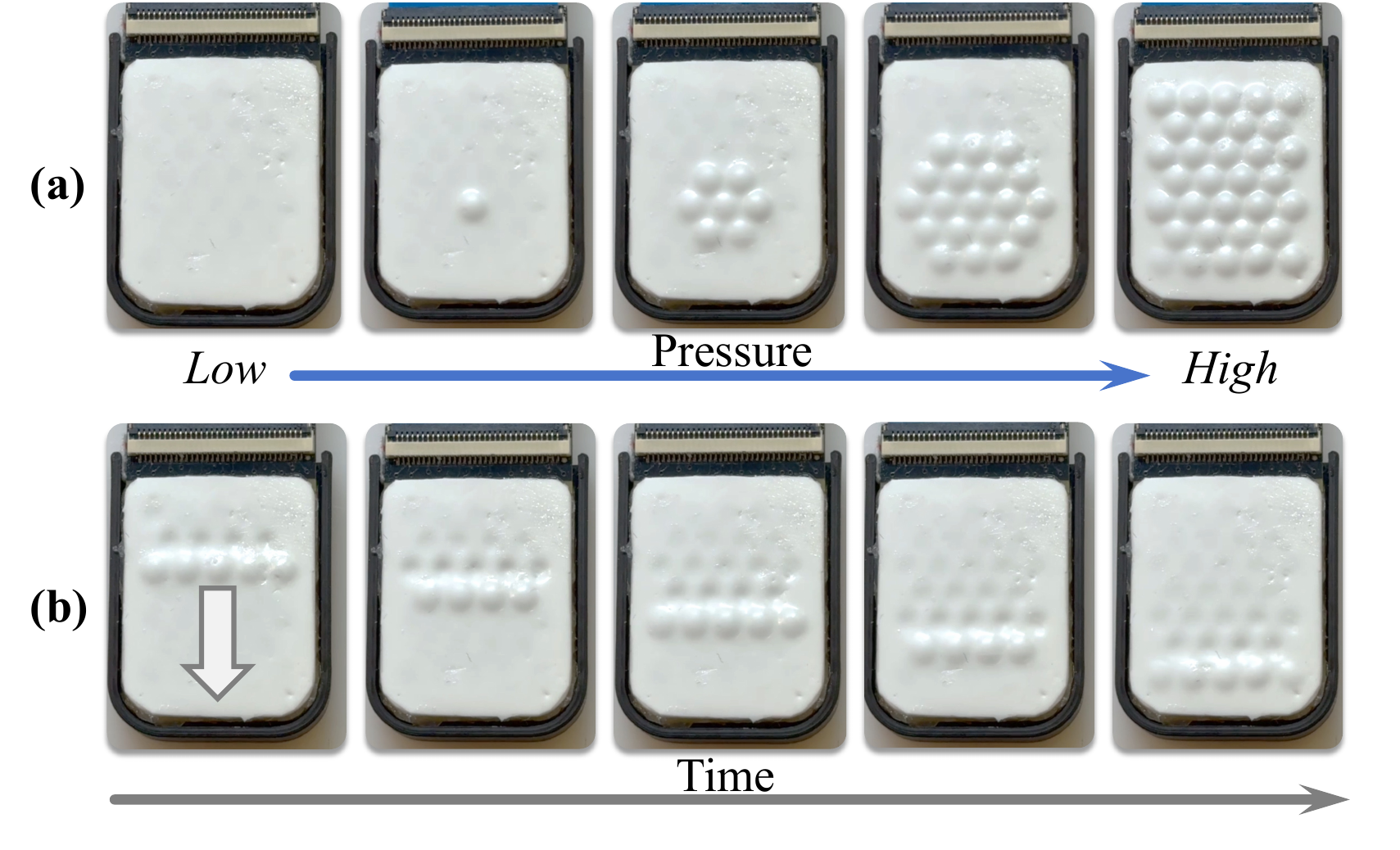}
  \vspace{-0.1in}
    \caption{\textbf{Dynamic demonstration of the \ours feedback module.} (a) Active area expansion in Pressure Mode under varying pressure levels. (b) Dynamic force profiles simulating forward-to-backward motion.}
  \label{fig:press}
    \vspace{-0.1in}
\end{figure}

\textbf{Pressure Mapping Mode:} This mode prioritizes the perception of force magnitude by converting intensity into contact area. Rather than a direct spatial map, the system monitors the normalized peak pressure $P_{max} \in [0, 1]$ from the sensor array. This value triggers a discrete, multi-level area expansion based on configurable thresholds (e.g., $T = \{0.1, 0.35, 0.7\}$):
\begin{equation}
    \mathcal{I}_{active} = \{k \mid k \in \text{Indices}(T_n) \text{ if } P_{max} \geq T_n\}.
\end{equation}
As $P_{max}$ surpasses each threshold, the active area expands radially from a single central point to 7 or 19 points. As shown in \cref{fig:press}(a), this stepwise expansion creates an intuitive \textit{tactile scale}, allowing the operator to gauge pressure levels through changes in the perceived contact area.

\subsection{Arm Pose Acquisition}
\label{sec:arm_retargt}
\ours{} is designed to be highly compatible with diverse teleoperation approaches for robot arms by simply adjusting its mechanical mounting adapters. In our experiments, we implement two primary methods for capturing wrist pose: utilizing the Forward Kinematics (FK) of the HOMIE exoskeleton~\cite{ben2025homie} and obtaining pose with HTC VIVE Trackers~\cite{vive2026}. This modular interfacing ensures that \ours{} can be readily integrated into various frameworks, including vision-based or VR-based systems, to meet specific user requirements.
\section{Hardware Evaluation}
\label{sec:eva}
In this section, we first evaluate the exoskeleton glove by characterizing its motion-tracking precision and robustness to EMI in \cref{subsec:glove_eva}. 
With reliable glove-based teleoperation established, we then characterize the tactile sensing performance of diverse robotic platforms in \cref{subsec:tac-eval} to inform hardware selection and validate our tactile mapping strategies.

\begin{figure*}[!ht]
  \centering
  \includegraphics[width=0.98\textwidth]{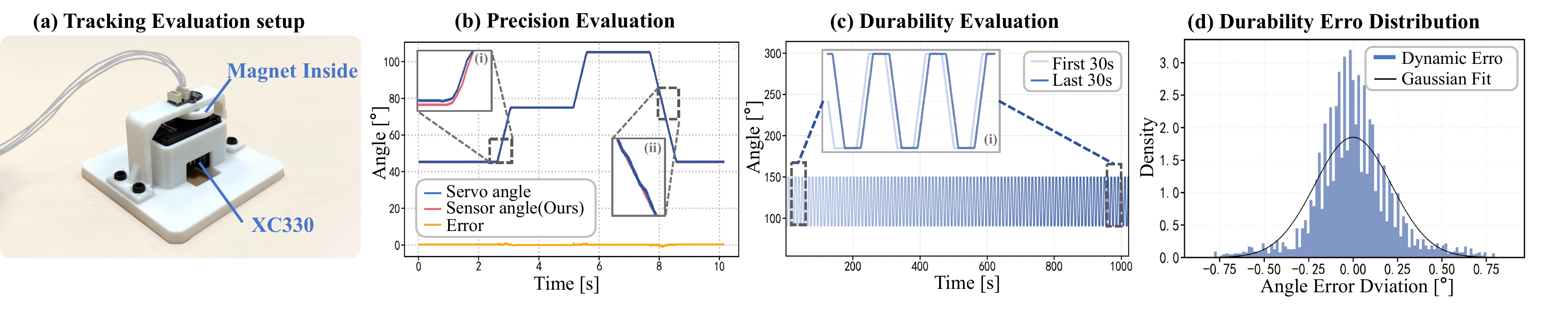}
  \vspace{-0.1in}
  \caption{\textbf{Exoskeleton glove tracking performance.} (a) Experimental setup for joint-tracking. (b) Real-time tracking results for discrete and continuous trajectories. (c) Long-term stability test over 1000s of operation. (d) Statistical distribution of instantaneous tracking errors with Gaussian fit.}
  \label{fig:glove_eva}
    \vspace{-0.15in}
\end{figure*}

\subsection{Glove Exoskeleton Evaluation}
\label{subsec:glove_eva}
\subsubsection{\textbf{High-Precision Joint Tracking}}
To quantify the joint-tracking accuracy of the \ours encoder, we use a Dynamixel XC330-M288-T servo ($0.088^\circ$ resolution) as a high-precision ground-truth reference, following established benchmarking protocols~\cite{shaw2023leaphand,zorin2025ruka}. As illustrated in \cref{fig:glove_eva}(a), a magnetic encoder module is mounted via a custom adapter to replicate its integration within the glove. The servo is commanded to execute a predefined angular trajectory ($45^\circ \rightarrow 75^\circ \rightarrow 105^\circ \rightarrow 45^\circ$), spanning a typical range of finger joint motion.

The tracking results in \cref{fig:glove_eva}(b) demonstrate that \ours achieves rapid response and high-fidelity tracking during both (i) discrete angle transitions and (ii) continuous motion. Across the entire sequence, the maximum tracking error remains within $\pm 0.35^\circ$, demonstrating sub-degree joint-tracking precision suitable for high-fidelity hand motion capture.

\subsubsection{\textbf{Long-Term Tracking Stability}}
Beyond instantaneous accuracy, sustained teleoperation also requires that such accuracy be maintained over extended periods. To evaluate long-term stability, the encoder is driven by the XC330 servo under prolonged reciprocating motion for $1000\,\text{s}$. As illustrated in \cref{fig:glove_eva}(c), a comparison between the initial and final $30\,\text{s}$ intervals shows highly consistent angular estimates, with an average discrepancy of only $\sim 0.02^\circ$, indicating negligible drift over extended operation.

Beyond evaluating start–end drift, we aim to characterize the distribution and temporal consistency of instantaneous tracking errors over the entire duration. We analyze the statistical characteristics of these errors, and the corresponding Gaussian fit is shown in \cref{fig:glove_eva}(d). The distribution is tightly centered at $0^\circ$, with a standard deviation of $\sigma = 0.215^\circ$, and the vast majority of samples fall within $\pm 0.8^\circ$. These results demonstrate stable and robust joint tracking during long-duration operation.

\subsubsection{\textbf{Strong EMI Resilience}}
While tracking accuracy is critical for sustained operation, teleoperation systems also need to remain reliable in electromagnetically complex environments, making robustness to electromagnetic interference (EMI) essential. To this end, we evaluate the EMI resilience of \ours by measuring motion-tracking stability while approaching an active PC chassis, a representative source of electromagnetic interference, and compare its performance against the commercial Manus glove~\cite{manus2026}. During the experiment, the operator gradually moves the index finger from a distance of $50\,\text{cm}$ to $5\,\text{cm}$ relative to the interference source.

\begin{figure}[!ht]
  \centering
  \includegraphics[width=0.48\textwidth]{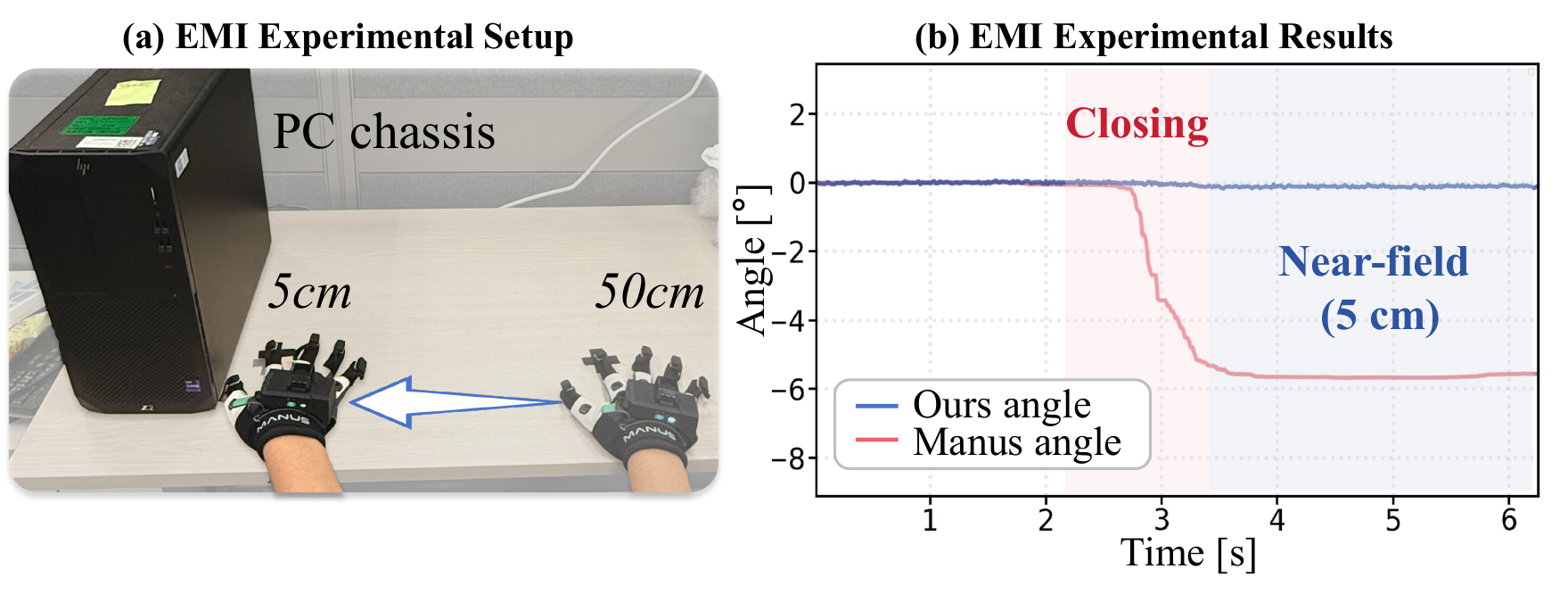}
  \vspace{-0.1in}
  \caption{\textbf{EMI resilience evaluation.} (a) Setup for near-field interference testing. (b) Angular stability comparison between the Manus glove and \ours{}.}
  \label{fig:EMI}
    \vspace{-0.1in}
\end{figure}

As shown in \cref{fig:EMI}(b), the Manus glove deviates noticeably during the approach phase (ii), reaching up to $5.69^\circ$. In contrast, \ours maintains stable tracking during approaching, with a maximum deviation of only $0.24^\circ$. These results highlight the superior EMI resilience of \ours, enabling reliable motion capture in electronics-intensive teleoperation scenarios. 

\begin{figure*}[!ht]
  \centering
  \includegraphics[width=0.98\textwidth]{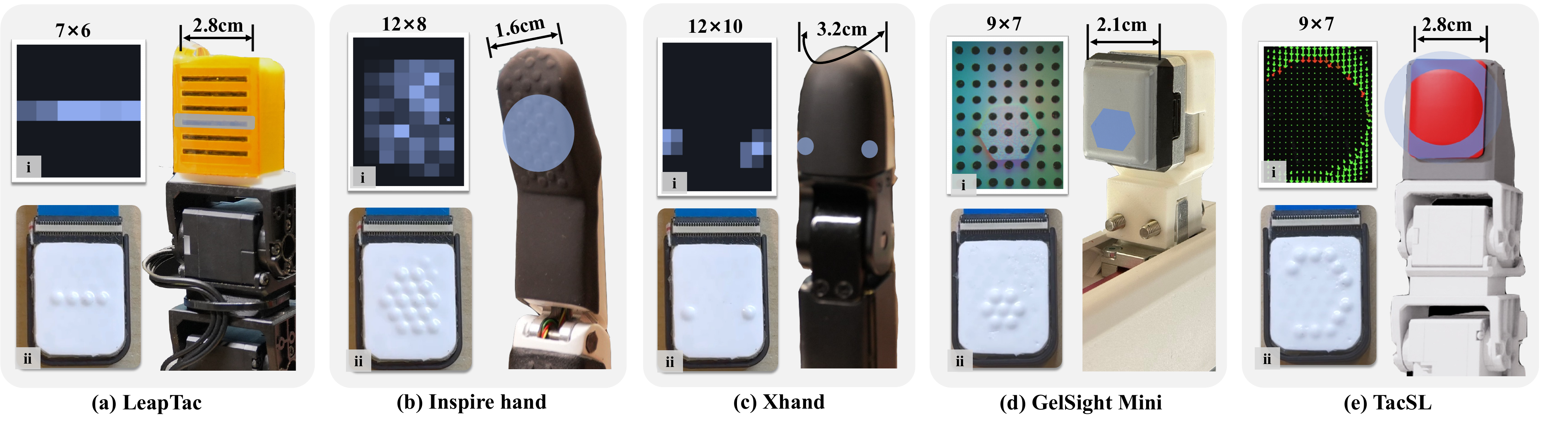}
    \caption{\textbf{Tactile sensing and shape mapping performance across various end-effectors.} 
      Comparison of raw sensing inputs (i) and physical rendering outputs on \ours{} (ii) across various robot end-effectors (right). 
      \ours effectively translates diverse sensor geometries into coherent tactile feedback: 
      (a) LeapTac: linear ridge; 
      (b) Inspire Hand: circular area; 
      (c) XHand: dual contact points; 
      (d) GelSight Mini and (e) TacSL: complex shapes.}
  \label{fig:sensor_eva}
  \vspace{-0.15in}
\end{figure*}

\subsection{Tactile Sensing Characterization}
\label{subsec:tac-eval}

Following the tactile mapping methodology established in \cref{sec:tactile_module}, practical implementation requires a characterization of the end-effector sensing capabilities. This evaluation is crucial not only for calibrating mapping parameters (e.g., gain and threshold) to ensure consistent feedback but also for selecting the most appropriate hardware platform for specific validation tasks. We characterize four distinct platforms: the custom resistive \textit{LeapTac} array~\cite{Huang20243DViTacLF} (integrated into LeapHand~\cite{shaw2023leaphand}), the commercial resistive Inspire Hand, the electromagnetic XHand, and the vision-based GelSight Mini.

\begin{table}[!ht]
    \centering
     \caption{\textbf{Comparative Analysis of Tactile Sensing Performance.} \textbf{Res:} Spatial Resolution. \textbf{Thres:} Minimum Force Threshold. \textbf{Var.} indicates sensitivity is tunable via gel stiffness. \textbf{EM:} Electromagnetic.}
       
    \label{tab:dexhand_sensors}
    \resizebox{.48\textwidth}{!}{
    \begin{tabular}{lccccc}
        \toprule
        \makecell[c]{\textbf{End-effector }} & 
        \makecell[c]{Sensing \\ Principle} & 
        \makecell[c]{Spatial \\ Res.} & 
        \makecell[c]{Sensing \\ Thres.(N)} & 
        \makecell[c]{Basic \\Shape} & 
        \makecell[c]{Complex\\Shape}  \\
        \midrule
        LeapTac& Resistive & $7 \times 6$ & 0.14 & \yes & \no \\
        Inspire& Resistive & $12 \times 8$ & 0.38 & \yes & \no \\
        XHand& EM-based & $12 \times 10$ & 0.08 & \half & \no \\
        GelSight Mini& Vision & $9 \times 7$ & Var. & \yes & \yes \\
        \bottomrule
    \end{tabular}}
    \vspace{-0.05in}
\end{table}

The quantitative evaluation results are summarized in \cref{tab:dexhand_sensors}. Based on these performance metrics, we assign specific platforms to corresponding validation experiments to maximize system efficacy. As indicated in the table, the Inspire Hand possesses a contiguous sensing area with sufficient spatial resolution ($12 \times 8$) to reliably distinguish basic geometric primitives (point, line, plane). Consequently, we select this platform to conduct the user studies detailed in \cref{sec:user}, as these tasks require stable spatial pattern reconstruction.

Conversely, the XHand features a fingertip-wrapping sensor. While this geometry limits the frontal contact area for planar differentiation ($\half$), it provides multi-directional sensing and exceptional sensitivity. Therefore, the XHand is assigned to the delicate Teleoperation Tasks in \cref{sec:teleop_perfo}, ensuring that even minimal interaction forces are captured and rendered.

Finally, while the GelSight Mini captures precise geometric shapes with resultant forces mapped to a $9 \times 7$ grid, and the custom LeapTac provides a cost-effective baseline, the integration of these diverse sensors validates the universality of our mapping framework. As illustrated in \cref{fig:sensor_eva}, by applying the proposed shape mapping mode, raw data from these varying modalities is successfully normalized into coherent tactile stimuli on the \ours glove.

\section{Teleoperation Experiments}

In this section, we conduct real-world teleoperation experiments using \ours on two different platforms: the Unitree G1 equipped with Inspire Hand, and the UR5e integrated with XHand. These experiments aim to demonstrate the capabilities and efficiency of \ours in contact-rich teleoperation tasks.
\subsection{User Study}
\label{sec:user}
To demonstrate how \ours empowers operators to ``feel what the robot feels'' and accurately discriminate the contact conditions, we conduct a user study with five trained participants. The study comprises two parts: contact shape discrimination and dynamic force direction discrimination, both conducted on the G1 robot with an Inspire Hand.

\subsubsection{Contact Shape Discrimination}
\label{sec:Contact_shape}
\mbox{}\par
\begin{figure}[!ht]
  \centering
  \includegraphics[width=0.49\textwidth]{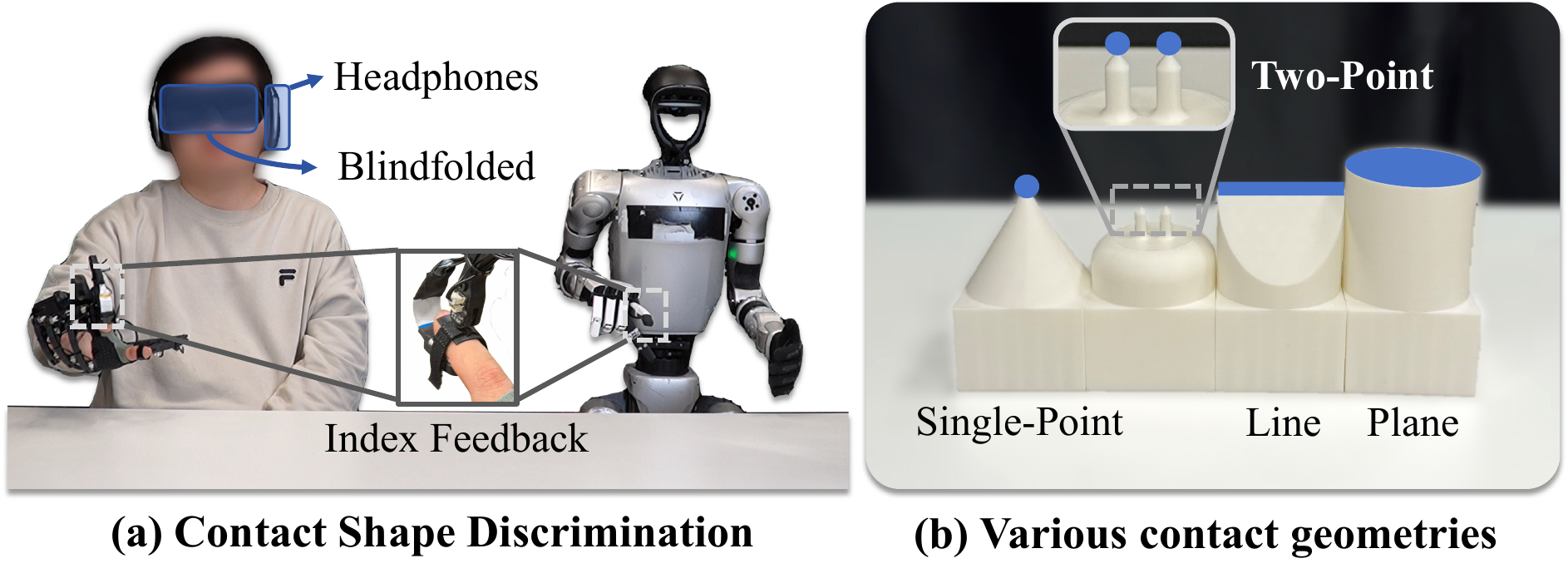}
    \caption{\textbf{User Study.} \textcolor{citecolor}{(a):} The operator, blindfolded and wearing headphones, identifies the shape of the object pressed against the index finger of G1 solely through tactile feedback. \textcolor{citecolor}{(b):} Four objects with different contact geometries: single-point, two-point, line, and plane.}
  \label{fig:user_study}
  \vspace{-0.1in}
\end{figure}

 \textbf{Setup:} As illustrated in \cref{fig:user_study}, the operator wears the \ours device and teleoperates G1 (equipped with the Inspire Hand), holding the arm raised with the index finger extended. An experimenter presses on the Inspire Hand's index finger using four objects with various geometries (specifically: single-point, two-point, line, and plane) for 5s. The operator, while blindfolded and wearing noise-canceling headphones, is required to identify the perceived objects based solely on the tactile feedback provided by the \ours device to their index finger. The typical tactile feedback patterns are shown in \cref{fig:sensor_eva}. A total of five participants are recruited, each performing 20 trials. The presentation order of the four objects is fully randomized, with each object being presented five times to each participant.

\begin{figure}[!ht]
  \centering
  \includegraphics[width=0.49\textwidth]{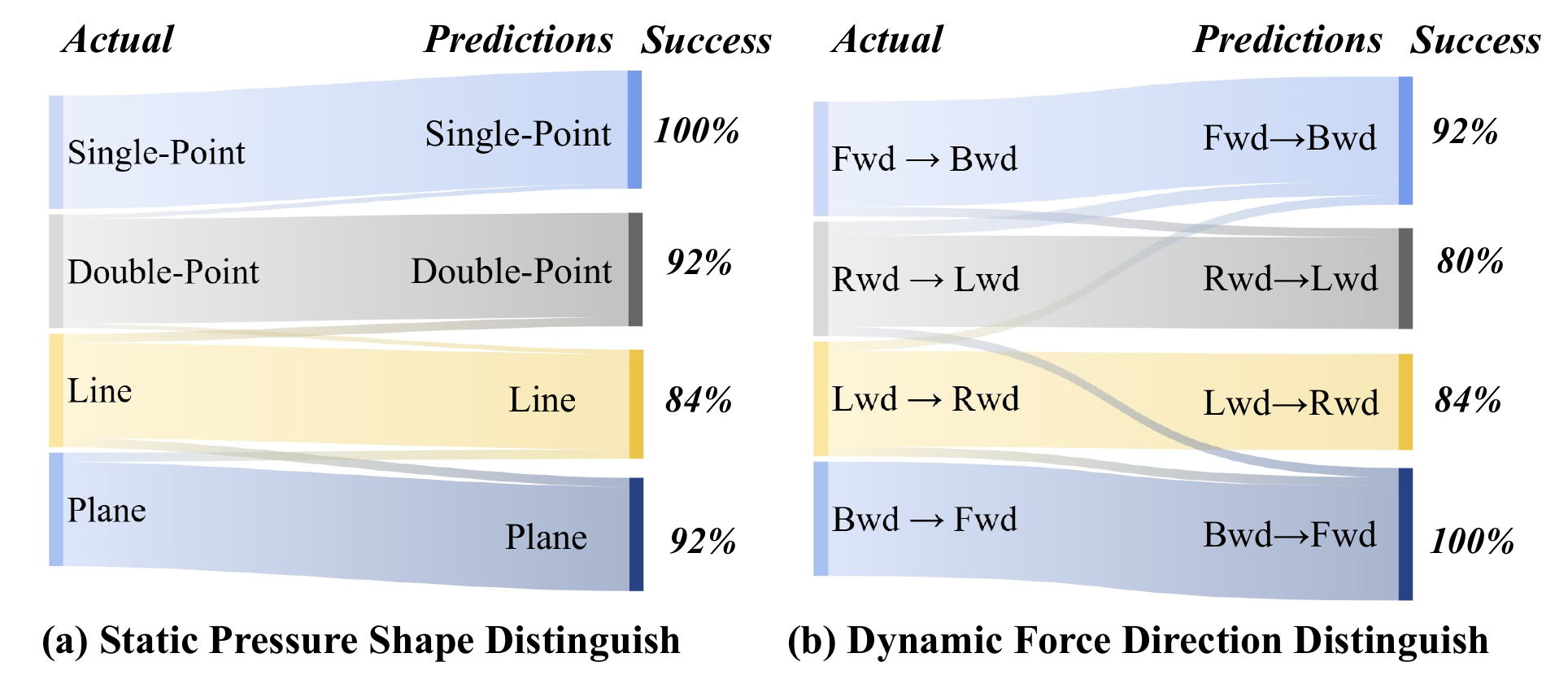}
    \caption{\textbf{Classification results of the user study.} 
      The Sankey diagrams visualize the mapping between actual stimuli and user predictions. 
      The width of the flows represents the number of trials.}
  \label{fig:user_results}
  \vspace{-0.15in}
\end{figure}

\textbf{Results:} The recognition performance for contact shape discrimination is visualized in \cref{fig:user_results}. The system demonstrates high fidelity in rendering spatial contact geometries particularly for \textbf{single-point} contacts, which are identified with a 100\% accuracy. Minor misidentifications primarily occur between geometrically similar categories; for instance, \textbf{line} contacts are occasionally perceived as \textbf{two-point} ($n=2$) or \textbf{plane} ($n=2$) contacts. Similarly, \textbf{two-point} and \textbf{plane} geometries achieve high recognition rates, with 23 correct identifications out of 25 trials each. The strong alignment between the actual and perceived geometries validates the effectiveness of the \ours device in conveying distinct contact patterns to the operator.

\subsubsection{Dynamic Force Direction Discrimination}
\label{sec:Dynamic_force}
\mbox{}\par \textbf{Setup:} The experimental setup follows the same protocol as the contact shape experiment. In this session, the experimenter applies a dynamic load by rolling a cylinder across the fingertip of the Inspire Hand for approximately 2s per trial. Four distinct motion directions are evaluated: \textbf{forward-to-backward}, \textbf{backward-to-forward}, \textbf{left-to-right}, and \textbf{right-to-left}. The dynamic evolution of these tactile patterns is illustrated in \textbf{\cref{fig:press}(b)}. Participants are required to determine the direction of the rolling force based solely on the dynamic tactile cues rendered by the \ours device. All other experimental conditions remain consistent with the contact shape discrimination study.

\textbf{Results:} As summarized in \cref{fig:user_results}, superior discriminability is observed for longitudinal motions, with participants achieving 100\% accuracy for \textbf{Bwd $\rightarrow$ Fwd} and 92\% (23/25) for \textbf{Fwd $\rightarrow$ Bwd}. Lateral recognition is slightly lower, with \textbf{R $\rightarrow$ L} and \textbf{L $\rightarrow$ R} correctly identified in 20 and 21 trials, respectively. This disparity stems from the physical aspect ratio of the \ours{} module; the narrower lateral dimension accommodates fewer sensing and actuation points compared to the longitudinal axis, resulting in reduced resolution for side-to-side movements. Nonetheless, the overall high recognition rates confirm that \ours{} effectively renders continuous directional cues, providing operators with intuitive motion perception.

\begin{figure*}[!ht]
\centering
  \includegraphics[width=0.98\textwidth]{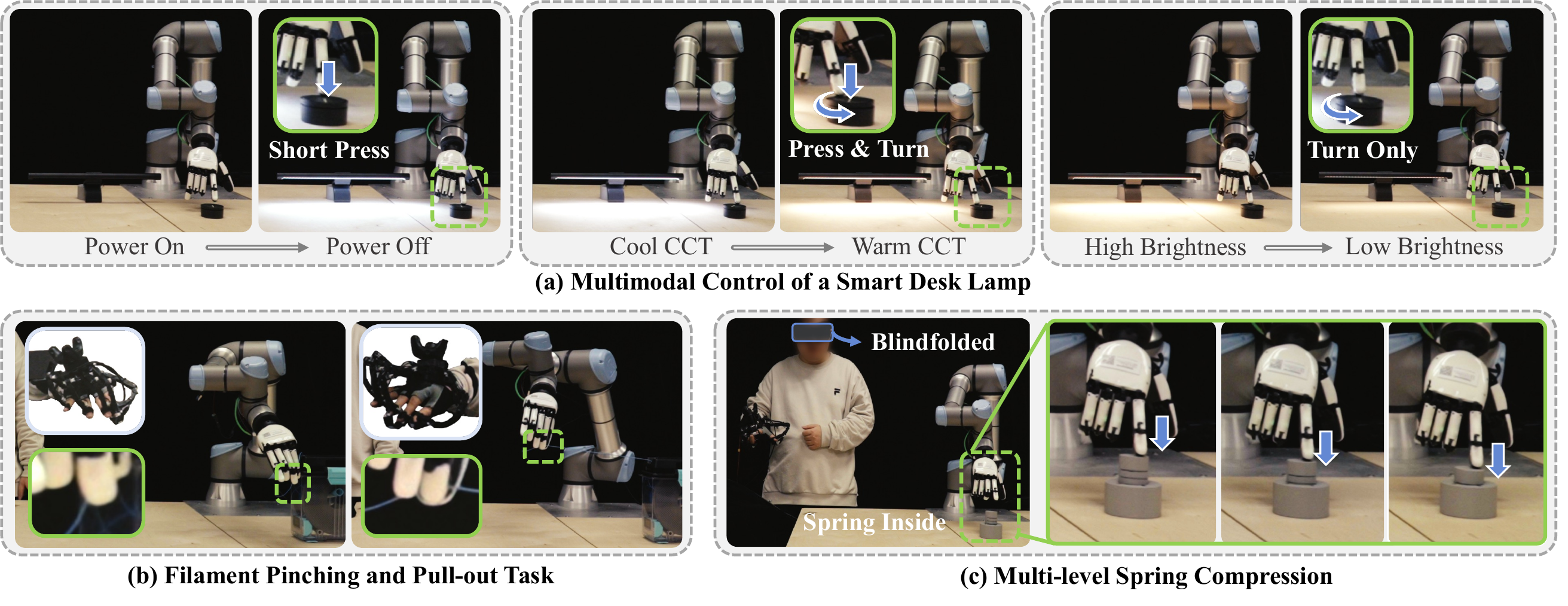}
  \vspace{-0.05in}
  \caption{\textbf{Real-world teleoperation tasks using \ours{}.} 
      (a) \textbf{Multimodal lamp control}: Hierarchical interaction via short press, press-and-turn, and subtle rotation. 
     (b) \textbf{Sequential filament pinching}: Distinguishing successful capture of a 1.75mm filament from empty pinches via localized tactile feedback. 
     (c) \textbf{Multi-level spring compression}: Maintaining three distinct pressure levels (0.5N, 3.5N, 7.5N) guided by rendered tactile feedback.}
  \label{fig:tele_tasks}
  \vspace{-0.15in}
\end{figure*}

\subsection{Teleoperation Performance}
\label{sec:teleop_perfo}
To evaluate \ours in complex contact-rich scenarios, we conduct three real-world teleoperation tasks using a UR5e arm with an XHand. These fine-motor tasks primarily involve the coordinated control of the index finger and thumb.

\subsubsection{\textbf{Multimodal Control of a Smart Desk Lamp}} 

\mbox{}\par \textbf{Tasks:} As shown in \cref{fig:tele_tasks}(a), this task requires hierarchical control of a smart lamp via a single rotary knob. To ensure precise force regulation, \ours is configured in \textbf{Press Mode}, mapping detected pressure to the feedback contact area. The task is decomposed into three stages:
\begin{itemize}[leftmargin=15pt, topsep=2pt, itemsep=1pt]
    \item \textbf{Stage 1: Power Toggling (Short Press)} – A \textit{short press} ($<$ 2s) toggles the power. Precise temporal control is essential as exceeding 2s triggers mode switching.
    \item \textbf{Stage 2: CCT Adjustment (Press \& Turn)} – The operator must maintain a steady downward force exceeding 3N while rotating the knob to adjust the color temperature. 
    \item \textbf{Stage 3: Brightness Adjustment (Turn Only)} – Rotation requires a subtle downward force within a narrow window (0.5N--3N). The force must be sufficient for friction but light enough to avoid accidentally triggering the ``press'' mode in Stage 2.
\end{itemize}
\vspace{-0.15in}
\mbox{}\par \textbf{Results:} Results are summarized in \cref{tab:performance_stages}. In \textbf{Stage 1}, although feedback slightly increases time (3.6s vs. 3.2s), the tactile ``click'' confirms actuation, ensuring a 100\% success rate. In \textbf{Stage 2}, array feedback allows operators to maintain stable pressure ($>$3N) by perceiving the full contact area. The benefit is most evident in \textbf{Stage 3}. Regulating force (0.5--3N) is difficult with single-point haptics, often leading to over-pressing ($>$3N). In contrast, \ours renders the \textit{incremental expansion} of the contact patch, providing intuitive confirmation at minimal force levels to safely engage the knob.

\begin{table}[!ht]
    \centering
    \caption{Results of multimodal desk lamp control.}
    \label{tab:performance_stages}
    \renewcommand{\arraystretch}{0.8} 
    \resizebox{.48\textwidth}{!}{
    \begin{tabular}{lcc}
        \toprule
        \textbf{Teleoperation System} & \textbf{Success Rate ($\uparrow$)} & \textbf{Average Time ($\downarrow$)} \\
        \midrule
        
        \ourrow \multicolumn{3}{l}{\textbf{Stage I:} Power Toggling (Short Press)} \\ 
        \noalign{\vskip 0.3mm} 
        \cdashline{1-3} 
        \noalign{\vskip 1.5mm} 
        HOMIE~\cite{ben2025homie} & 12/15 & 6.1s \\
        \ours w/o feedback & 14/15 & \textbf{3.2s} \\
        \textbf{\ours w/ feedback} & \textbf{15/15} & 3.6s \\
        \midrule
        
        \ourrow \multicolumn{3}{l}{\textbf{Stage II:} CCT Adjustment (Press \& Turn)} \\ 
        \noalign{\vskip 0.3mm}
        \cdashline{1-3}
        \noalign{\vskip 1.5mm} 
        HOMIE~\cite{ben2025homie} & 9/15 & 23.5s \\
        \ours w/o feedback & 12/15 & 14.8s \\
        \textbf{\ours w/ feedback} & \textbf{14/15} & \textbf{8.4s} \\
        \midrule
        
        \ourrow \multicolumn{3}{l}{\textbf{Stage III:} Brightness Adjustment (Turn Only)} \\ 
        \noalign{\vskip 0.3mm}
        \cdashline{1-3}
        \noalign{\vskip 1.5mm} 
        HOMIE~\cite{ben2025homie} & 1/15 & 31.3s \\
        \ours w/o feedback & 4/15 & 26.2s \\
        \textbf{\ours w/ feedback} & \textbf{12/15} & \textbf{15.7s} \\
        
        \bottomrule
    \end{tabular}}
    \vspace{-0.05in}
\end{table}

\subsubsection{\textbf{Filament Pinching and Pull-out}} 

\mbox{}\par \textbf{Tasks:} As shown in \cref{fig:tele_tasks}(b), this task requires the operator to pinch and extract a 1.75mm diameter 3D printing filament in two consecutive trials. Each trial requires a stable pull exceeding 20cm in length. With \ours in \textbf{Shape Mode}, the operator must distinguish between an \textit{empty pinch} (diffuse tactile feedback area) and a \textit{successful capture} (narrow tactile feedback area), confirming the filament is securely centered before initiating the long pull.
\mbox{}\par \textbf{Results:} The results are summarized in \cref{tab:filament_performance}. In the absence of feedback, operators frequently fail either midway through the pulling motion or during a second attempt due to tactile ambiguity, often initiating the pull before the filament is securely centered. In contrast, \ours provides distinct geometric differentiation. The transition to a sharp, narrow tactile feedback area serves as an intuitive ``trigger signal,'' ensuring secure engagement for the long-distance pull. The precision feedback from \ours allows for an 86.7\% (13/15) success rate and reduces completion time to 22.4s.

\begin{table}[!ht]
    \centering
    \caption{Results of sequential filament pinching (2 consecutive trials).}
    \label{tab:filament_performance}
    \renewcommand{\arraystretch}{1.0} 
    \resizebox{.44\textwidth}{!}{
    \begin{tabular}{lcc}
        \toprule
        \textbf{Teleoperation System} & \textbf{Success Rate ($\uparrow$)} & \textbf{Average Time ($\downarrow$)} \\
        \midrule

        HOMIE~\cite{ben2025homie} & 5/15 & 42.8s \\
        \ours w/o feedback & 7/15 & 35.6s \\
        \textbf{\ours w/ feedback} & \textbf{13/15} & \textbf{22.4s} \\
        
        \bottomrule
    \end{tabular}}
\end{table}

\subsubsection{\textbf{Multi-level Spring Compression}} 

\mbox{}\par \textbf{Tasks:} As shown in \cref{fig:tele_tasks}(c), this task evaluates the precise control of force magnitude using a spring-loaded button. \ours is configured in \textbf{Press Mode}. Operators must sequentially reach and maintain three pressure levels for $>$2s: \textit{Initial Contact} (0.5N), \textit{Half-compressed} (3.5N), and \textit{Fully-compressed} (7.5N). Tests are conducted under two conditions, blindfolded and with visual feedback, to assess the system's ability to substitute visual cues with tactile feedback.
\mbox{}\par \textbf{Results:} As shown in \cref{tab:spring_performance}, baseline systems under blindfolded conditions yield success rates $\leq$ 20\%, with operators noting these few successes are purely stochastic and lack repeatability. Without tactile feedback, operators struggle to distinguish pressing force, resulting in unreliable proprioceptive cues. In contrast, \ours enables the operators to achieve an 86.7\% success rate by rendering the incremental expansion of the contact patch as intuitive feedback. This allows operators to ``lock'' onto specific force levels with high confidence. Supplementing visual feedback with tactile feedback via \ours increases the success rate from 12/15 to 15/15, confirming that \ours provides a more immediate and precise signal for force maintenance than visual estimation of spring displacement.
\begin{table}[!ht]
    \centering
    \caption{Success rate of multi-level spring compression.}
    \label{tab:spring_performance}
    \renewcommand{\arraystretch}{0.9} 
    \resizebox{.45\textwidth}{!}{
    \begin{tabular}{lcc}
        \toprule
        \textbf{Teleoperation System} & \textbf{Blindfolded ($\uparrow$)} & \textbf{With Visual ($\uparrow$)} \\
        \midrule
        HOMIE~\cite{ben2025homie} & 2/15 & 10/15 \\
        \ours w/o feedback & 3/15 & 12/15 \\
        \textbf{\ours w/ feedback} & \textbf{13/15} & \textbf{15/15} \\
        \bottomrule
    \end{tabular}}
    \vspace{-0.15in}
\end{table}

\subsection{Imitation Learning}
\label{sec:IL}
\begin{figure}[!ht]
  \centering
  \includegraphics[width=0.45\textwidth]{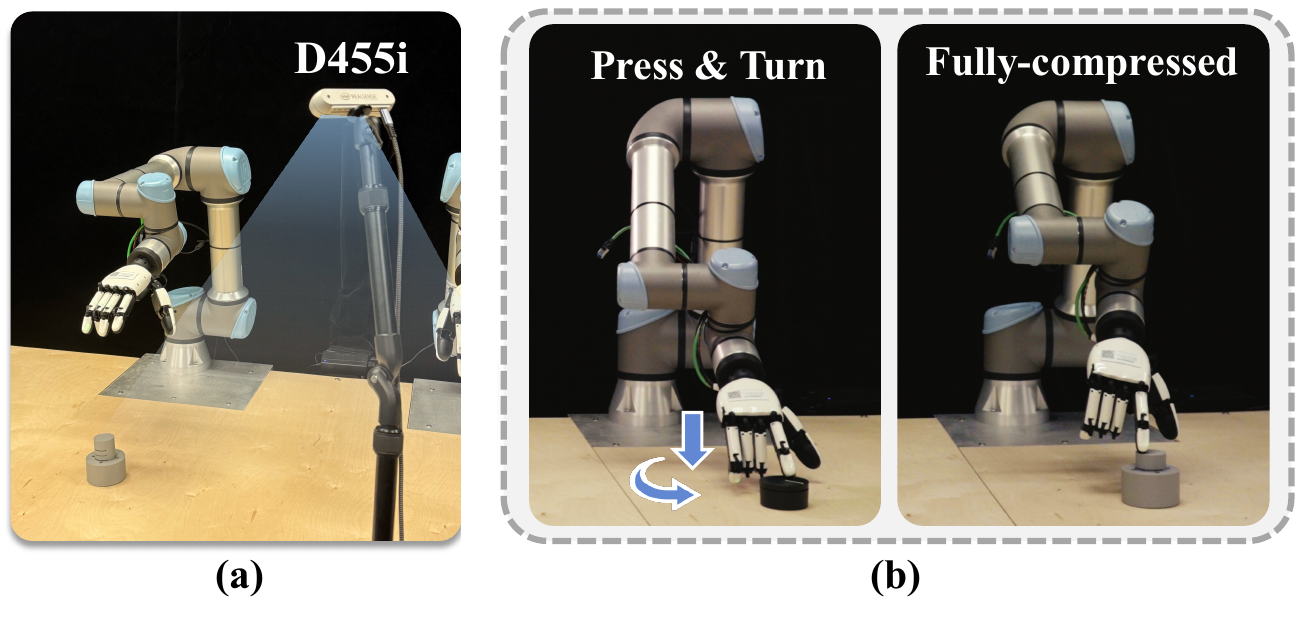}
  \vspace{-0.1in}
    \caption{\textbf{Imitation learning.} 
      (a) Experimental setup for real-world tasks.
     (b) Task illustration: CCT Adjustment and Fully Spring Compression.}
  \label{fig:IL}
    \vspace{-0.1in}
\end{figure}

We train a manipulation policy via imitation learning to verify the usability of data collected through \ours teleoperation. ManiFlow~\cite{maniflow} is employed as the policy. ManiFlow takes point clouds and the robot state as input and outputs actions. The actions are defined as the end-effector pose of the robot arm and the joint positions of the robot hand. As illustrated in \cref{fig:IL}(a), we use a RealSense D455 camera to capture scene point clouds. Tactile information is included as part of the robot state input to the policy. We evaluate the learned policy on two representative tasks: CCT Adjustment and Full Spring Compression, as shown in \cref{fig:IL}(b). The objects are randomly placed within a \(0.1 \times 0.1\,\mathrm{m}\) area. The policy trained on data collected via \ours achieves success rates of up to 15/15 and 13/15 on the two tasks, demonstrating the effectiveness of \ours data collection.

\section{Conclusion and Limitations}
In this paper, we introduce \ours, a low-cost glove system integrating precise motion capture with high-resolution tactile feedback. By leveraging 21-DoF non-contact 3-axis magnetic sensors, our system achieves sub-degree precision and interference-resistant joint tracking. To bridge the sensory gap, we incorporate compact EEOP arrays that render tactile cues via dual mapping strategies, effectively encoding contact shapes and pressure magnitudes. Real-world experiments validate that \ours{} empowers operators to perform delicate, contact-rich tasks with superior situational awareness compared to baselines. Ultimately, this approach significantly increases the reliability of demonstration data collection for learning-based manipulation.

\textbf{Limitations}
Despite its capabilities, \ours{} has limitations. First, while our EEOP modules provide high-resolution pressure and geometry cues, they lack kinesthetic feedback to physically restrict finger motion. Furthermore, although dynamic shear can be approximated through temporal modulation, the system cannot render static shear forces necessary for detecting steady-state slip. Second, the manual assembly of fluidic circuits introduces variability and potential durability concerns. Future work will focus on hybrid actuation and standardized micro-fabrication to address these issues.

\section*{Acknowledgments}
This work is supported by the Shanghai Artificial Intelligence Laboratory. We thank Zhongwei Xiang, Jingping Wang, Huayi Wang, and Xiao Chen for their helpful assistance and valuable discussions. We are grateful to Shenghan Zhang and Quanli Xuan for their support in hardware development, and Yanming Shao for coordinating the shared use of the Xhand.

\bibliographystyle{plainnat}
\bibliography{references}

\begin{thebibliography}{43}
\providecommand{\natexlab}[1]{#1}
\providecommand{\url}[1]{\texttt{#1}}
\expandafter\ifx\csname urlstyle\endcsname\relax
  \providecommand{\doi}[1]{doi: #1}\else
  \providecommand{\doi}{doi: \begingroup \urlstyle{rm}\Url}\fi

\bibitem[Ben et~al.(2025)Ben, Jia, Zeng, Dong, Lin, and Pang]{ben2025homie}
Qingwei Ben, Feiyu Jia, Jia Zeng, Junting Dong, Dahua Lin, and Jiangmiao Pang.
\newblock Homie: Humanoid loco-manipulation with isomorphic exoskeleton cockpit.
\newblock In \emph{Robotics Science and Systems (RSS)}, 2025.

\bibitem[Black et~al.(2024)Black, Brown, Driess, Esmail, Equi, Finn, Fusai, Groom, Hausman, Ichter, et~al.]{pi0}
Kevin Black, Noah Brown, Danny Driess, Adnan Esmail, Michael Equi, Chelsea Finn, Niccolo Fusai, Lachy Groom, Karol Hausman, Brian Ichter, et~al.
\newblock $\pi_0$: A vision-language-action flow model for general robot control.
\newblock \emph{arXiv preprint arXiv:2410.24164}, 2024.

\bibitem[Chen et~al.(2026)Chen, Ni, Lou, Shen, Wei, Liu, and Song]{forceglove}
Dapeng Chen, Haojun Ni, Juncheng Lou, Lianshun Shen, Zhong Wei, Jia Liu, and Aiguo Song.
\newblock Design and evaluation of a compact force feedback glove for virtual reality applications.
\newblock \emph{Sensors and Actuators A: Physical}, 2026.

\bibitem[Cheng et~al.(2024)Cheng, Li, Yang, Yang, and Wang]{cheng2024open}
Xuxin Cheng, Jialong Li, Shiqi Yang, Ge~Yang, and Xiaolong Wang.
\newblock Open-television: Teleoperation with immersive active visual feedback.
\newblock In \emph{Conference on Robot Learning (CoRL)}, 2024.

\bibitem[Chi et~al.(2025)Chi, Xu, Feng, Cousineau, Du, Burchfiel, Tedrake, and Song]{dp}
Cheng Chi, Zhenjia Xu, Siyuan Feng, Eric Cousineau, Yilun Du, Benjamin Burchfiel, Russ Tedrake, and Shuran Song.
\newblock Diffusion policy: Visuomotor policy learning via action diffusion.
\newblock \emph{The International Journal of Robotics Research (IJRR)}, 2025.

\bibitem[Ding et~al.(2025)Ding, Qin, Zhu, Jia, Yang, Yang, Qi, and Wang]{ding2024bunny}
Runyu Ding, Yuzhe Qin, Jiyue Zhu, Chengzhe Jia, Shiqi Yang, Ruihan Yang, Xiaojuan Qi, and Xiaolong Wang.
\newblock Bunny-visionpro: Real-time bimanual dexterous teleoperation for imitation learning.
\newblock In \emph{International Conference on Intelligent Robots and Systems (IROS)}, 2025.

\bibitem[Fang et~al.(2025)Fang, Romero, Xie, Hu, Huang, Alvarez, Kim, Margolis, Anbarasu, Tomizuka, et~al.]{fang2025dexop}
Hao-Shu Fang, Branden Romero, Yichen Xie, Arthur Hu, Bo-Ruei Huang, Juan Alvarez, Matthew Kim, Gabriel Margolis, Kavya Anbarasu, Masayoshi Tomizuka, et~al.
\newblock Dexop: A device for robotic transfer of dexterous human manipulation.
\newblock \emph{arXiv preprint arXiv:2509.04441}, 2025.

\bibitem[Feng et~al.(2025)Feng, Fang, He, Chen, Wang, He, Liu, and Lu]{feng2025learning}
Ying Feng, Hongjie Fang, Yinong He, Jingjing Chen, Chenxi Wang, Zihao He, Ruonan Liu, and Cewu Lu.
\newblock Learning dexterous manipulation with quantized hand state.
\newblock \emph{arXiv preprint arXiv:2509.17450}, 2025.

\bibitem[Handa et~al.(2020)Handa, Van~Wyk, Yang, Liang, Chao, Wan, Birchfield, Ratliff, and Fox]{DexPilot}
Ankur Handa, Karl Van~Wyk, Wei Yang, Jacky Liang, Yu-Wei Chao, Qian Wan, Stan Birchfield, Nathan Ratliff, and Dieter Fox.
\newblock Dexpilot: Vision-based teleoperation of dexterous robotic hand-arm system.
\newblock In \emph{International Conference on Robotics and Automation (ICRA)}, 2020.

\bibitem[HaptX(2026)]{haptx2026}
HaptX.
\newblock Haptx official website, 2026.
\newblock URL \url{https://haptx.com/}.

\bibitem[Harrison et~al.(2024)Harrison, Jester, Mouli, Fratini, and Jabran]{Harrison2024imu}
A.~Harrison, A.~Jester, S.~Mouli, A.~Fratini, and A.~Jabran.
\newblock Systematic evaluation of {IMU} sensors for application in smart glove system for remote monitoring of hand differences.
\newblock \emph{Sensors}, 2024.

\bibitem[Heng et~al.(2025)Heng, Geng, Zhang, Abbeel, and Malik]{heng2025vitacformer}
Liang Heng, Haoran Geng, Kaifeng Zhang, Pieter Abbeel, and Jitendra Malik.
\newblock Vitacformer: Learning cross-modal representation for visuo-tactile dexterous manipulation.
\newblock \emph{arXiv preprint arXiv:2506.15953}, 2025.

\bibitem[Huang et~al.(2024)Huang, Wang, Yang, Luo, and Li]{Huang20243DViTacLF}
Binghao Huang, Yixuan Wang, Xinyi Yang, Yiyue Luo, and Yunzhu Li.
\newblock 3d-vitac: Learning fine-grained manipulation with visuo-tactile sensing.
\newblock In \emph{Conference on Robot Learning (CoRL)}, 2024.

\bibitem[Jiang et~al.(2024)Jiang, Fan, Xie, Kuang, Zhang, and Fan]{JiangElec}
Chutian Jiang, Yinan Fan, Junan Xie, Emily Kuang, Kaihao Zhang, and Mingming Fan.
\newblock Designing unobtrusive modulated electrotactile feedback on fingertip edge to assist blind and low vision (blv) people in comprehending charts.
\newblock In \emph{Conference on Human Factors in Computing Systems}, 2024.

\bibitem[Li et~al.(2020)Li, Jiang, Ruppel, Liang, Ma, Hendrich, Sun, and Zhang]{imutele}
Shuang Li, Jiaxi Jiang, Philipp Ruppel, Hongzhuo Liang, Xiaojian Ma, Norman Hendrich, Fuchun Sun, and Jianwei Zhang.
\newblock A mobile robot hand-arm teleoperation system by vision and imu.
\newblock In \emph{International Conference on Intelligent Robots and Systems (IROS)}, 2020.

\bibitem[Lin et~al.(2025)Lin, Zhang, Li, Qi, Yi, Levine, and Malik]{lin2025learning}
Toru Lin, Yu~Zhang, Qiyang Li, Haozhi Qi, Brent Yi, Sergey Levine, and Jitendra Malik.
\newblock Learning visuotactile skills with two multifingered hands.
\newblock In \emph{International Conference on Robotics and Automation (ICRA)}, 2025.

\bibitem[Lu et~al.(2025)Lu, Cheng, Li, Yang, Ji, Yuan, Yang, Yi, and Wang]{lu2025mobile}
Chenhao Lu, Xuxin Cheng, Jialong Li, Shiqi Yang, Mazeyu Ji, Chengjing Yuan, Ge~Yang, Sha Yi, and Xiaolong Wang.
\newblock Mobile-television: Predictive motion priors for humanoid whole-body control.
\newblock In \emph{International Conference on Robotics and Automation (ICRA)}, 2025.

\bibitem[MetaGloves(2026)]{manus2026}
MetaGloves.
\newblock Metagloves official website, 2026.
\newblock URL \url{https://manus-meta.com/}.

\bibitem[Qin et~al.(2024)Qin, Yang, Huang, Wyk, Su, Wang, Chao, and Fox]{anyteleop}
Yuzhe Qin, Wei Yang, Binghao Huang, Karl~Van Wyk, Hao Su, Xiaolong Wang, Yu-Wei Chao, and Dieter Fox.
\newblock Anyteleop: A general vision-based dexterous robot arm-hand teleoperation system.
\newblock In \emph{Robotics Science and Systems (RSS)}, 2024.

\bibitem[Romero et~al.(2024)Romero, Fang, Agrawal, and Adelson]{romero2024eyesight}
Branden Romero, Hao-Shu Fang, Pulkit Agrawal, and Edward Adelson.
\newblock Eyesight hand: Design of a fully-actuated dexterous robot hand with integrated vision-based tactile sensors and compliant actuation.
\newblock In \emph{International Conference on Intelligent Robots and Systems (IROS)}, 2024.

\bibitem[Shaw et~al.(2023)Shaw, Agarwal, and Pathak]{shaw2023leaphand}
Kenneth Shaw, Ananye Agarwal, and Deepak Pathak.
\newblock Leap hand: Low-cost, efficient, and anthropomorphic hand for robot learning.
\newblock In \emph{Robotics Science and Systems (RSS)}, 2023.

\bibitem[Shen et~al.(2023)Shen, Rae-Grant, Mullenbach, Harrison, and Shultz]{FluidReality}
Vivian Shen, Tucker Rae-Grant, Joe Mullenbach, Chris Harrison, and Craig Shultz.
\newblock Fluid reality: High-resolution, untethered haptic gloves using electroosmotic pump arrays.
\newblock In \emph{ACM Symposium on User Interface Software and Technology}, 2023.

\bibitem[Shultz and Harrison(2023)]{FlatPanel}
Craig Shultz and Chris Harrison.
\newblock Flat panel haptics: Embedded electroosmotic pumps for scalable shape displays.
\newblock In \emph{Conference on Human Factors in Computing Systems}, 2023.

\bibitem[Si and Yuan(2022)]{si2022taxim}
Zilin Si and Wenzhen Yuan.
\newblock Taxim: An example-based simulation model for gelsight tactile sensors.
\newblock \emph{Robotics and Automation Letters (RA-L)}, 2022.

\bibitem[Si et~al.(2024)Si, Zhang, Ben, Romero, Xian, Liu, and Gan]{si2024difftactile}
Zilin Si, Gu~Zhang, Qingwei Ben, Branden Romero, Zhou Xian, Chao Liu, and Chuang Gan.
\newblock Difftactile: A physics-based differentiable tactile simulator for contact-rich robotic manipulation.
\newblock In \emph{International Conference on Learning Representations (ICLR)}, 2024.

\bibitem[Sivakumar et~al.(2022)Sivakumar, Shaw, and Pathak]{sivakumar2022robotic}
Aravind Sivakumar, Kenneth Shaw, and Deepak Pathak.
\newblock Robotic telekinesis: Learning a robotic hand imitator by watching humans on youtube.
\newblock In \emph{Robotics Science and Systems (RSS)}, 2022.

\bibitem[Team(2025)]{generalist2025gen0}
Generalist~AI Team.
\newblock Gen-0: Embodied foundation models that scale with physical interaction.
\newblock \emph{Generalist AI Blog}, 2025.
\newblock https://generalistai.com/blog/nov-04-2025-GEN-0.

\bibitem[Tracker(2026)]{vive2026}
VIVE Tracker.
\newblock Vive official website, 2026.
\newblock URL \url{https://vive.com/}.

\bibitem[Vizcay et~al.(2021)Vizcay, Kourtesis, Argelaguet, Pacchierotti, and Marchal]{VizcayElec}
Sebastian Vizcay, Panagiotis Kourtesis, Ferran Argelaguet, Claudio Pacchierotti, and Maud Marchal.
\newblock Electrotactile feedback for enhancing contact information in virtual reality.
\newblock \emph{International Conference on Artificial Reality and Telexistence and Eurographics Symposium on Virtual Environments}, 2021.

\bibitem[Wang et~al.(2023)Wang, Fan, Sun, Zhang, Fei-Fei, Xu, Zhu, and Anandkumar]{wang2023mimicplay}
Chen Wang, Linxi Fan, Jiankai Sun, Ruohan Zhang, Li~Fei-Fei, Danfei Xu, Yuke Zhu, and Anima Anandkumar.
\newblock Mimicplay: Long-horizon imitation learning by watching human play.
\newblock In \emph{Conference on Robot Learning (CoRL)}, 2023.

\bibitem[Wang et~al.(2024)Wang, Shi, Wang, Zhang, Fei-Fei, and Liu]{wang2024dexcap}
Chen Wang, Haochen Shi, Weizhuo Wang, Ruohan Zhang, Li~Fei-Fei, and C~Karen Liu.
\newblock Dexcap: Scalable and portable mocap data collection system for dexterous manipulation.
\newblock In \emph{Robotics Science and Systems (RSS)}, 2024.

\bibitem[Wen et~al.(2025)Wen, Chen, Cui, Du, Gou, Han, Huang, Lei, Li, Li, et~al.]{wen2025gr}
Ruoshi Wen, Guangzeng Chen, Zhongren Cui, Min Du, Yang Gou, Zhigang Han, Liqun Huang, Mingyu Lei, Yunfei Li, Zhuohang Li, et~al.
\newblock Gr-dexter technical report.
\newblock \emph{arXiv preprint arXiv:2512.24210}, 2025.

\bibitem[Wu et~al.(2024)Wu, Shentu, Yi, Lin, and Abbeel]{wu2024gello}
Philipp Wu, Yide Shentu, Zhongke Yi, Xingyu Lin, and Pieter Abbeel.
\newblock Gello: A general, low-cost, and intuitive teleoperation framework for robot manipulators.
\newblock In \emph{International Conference on Intelligent Robots and Systems (IROS)}, 2024.

\bibitem[Xu et~al.(2025)Xu, Zhang, Hou, Xu, Fan, Veloso, and Song]{xu2025dexumi}
Mengda Xu, Han Zhang, Yifan Hou, Zhenjia Xu, Linxi Fan, Manuela Veloso, and Shuran Song.
\newblock Dexumi: Using human hand as the universal manipulation interface for dexterous manipulation.
\newblock In \emph{Conference on Robot Learning (CoRL)}, 2025.

\bibitem[Yan et~al.(2025)Yan, Zhu, Deng, Yang, Qiu, Cheng, Memmel, Krishna, Goyal, Wang, et~al.]{maniflow}
Ge~Yan, Jiyue Zhu, Yuquan Deng, Shiqi Yang, Ri-Zhao Qiu, Xuxin Cheng, Marius Memmel, Ranjay Krishna, Ankit Goyal, Xiaolong Wang, et~al.
\newblock Maniflow: A general robot manipulation policy via consistency flow training.
\newblock In \emph{Conference on Robot Learning (CoRL)}, 2025.

\bibitem[Yang(2024)]{yang2025ace}
Shiqi Yang.
\newblock Ace: A cross-platform visual-exoskeleton system for low-cost dexterous teleoperation.
\newblock In \emph{Conference on Robot Learning (CoRL)}, 2024.

\bibitem[Yu et~al.(2025)Yu, He, Wei, Wang, Li, Wang, Lu, Yue, Teng, Wang, Lin, Mi, Lu, and Yao]{MorphingSkin}
Tianyu Yu, Peisheng He, Bob~Tianqi Wei, Chenyuheng Wang, Xueqing Li, Xuezhu Wang, Yao Lu, Wei Yue, Megan Teng, Zihan Wang, Liwei Lin, Haipeng Mi, Qi~Lu, and Lining Yao.
\newblock Morphingskin: A skin-like platform that integrates multimodal hydraulic actuators based on flexible electroosmotic pumps.
\newblock In \emph{ACM Symposium on User Interface Software and Technology}, 2025.

\bibitem[Yuan et~al.(2017)Yuan, Dong, and Adelson]{yuan2017gelsight}
Wenzhen Yuan, Siyuan Dong, and Edward~H Adelson.
\newblock Gelsight: High-resolution robot tactile sensors for estimating geometry and force.
\newblock \emph{Sensors}, 2017.

\bibitem[Ze et~al.(2024)Ze, Zhang, Zhang, Hu, Wang, and Xu]{dp3}
Yanjie Ze, Gu~Zhang, Kangning Zhang, Chenyuan Hu, Muhan Wang, and Huazhe Xu.
\newblock 3d diffusion policy: Generalizable visuomotor policy learning via simple 3d representations.
\newblock In \emph{Robotics Science and Systems (RSS)}, 2024.

\bibitem[Zhang et~al.(2025)Zhang, Hu, Yuan, and Xu]{zhang2025do}
Han Zhang, Songbo Hu, Zhecheng Yuan, and Huazhe Xu.
\newblock Doglove: Dexterous manipulation with a low-cost open-source haptic force feedback glove.
\newblock In \emph{Robotics Science and Systems (RSS)}, 2025.

\bibitem[Zhao et~al.(2023)Zhao, Kumar, Levine, and Finn]{act}
Tony~Z Zhao, Vikash Kumar, Sergey Levine, and Chelsea Finn.
\newblock Learning fine-grained bimanual manipulation with low-cost hardware.
\newblock In \emph{Robotics Science and Systems (RSS)}, 2023.

\bibitem[Zhu et~al.(2021)Zhu, Stuttaford-Fowler, Fahmy, Li, and Sienz]{flexsensor}
Shuo Zhu, Angus Stuttaford-Fowler, Ashraf Fahmy, Chunxu Li, and Johann Sienz.
\newblock Development of a low-cost data glove using flex sensors for the robot hand teleoperation.
\newblock In \emph{Intelligent Manufacturing Technology (ISRIMT)}, 2021.

\bibitem[Zorin et~al.(2025)Zorin, Guzey, Yan, Iyer, Kondrich, Bhattasali, and Pinto]{zorin2025ruka}
Anya Zorin, Irmak Guzey, Billy Yan, Aadhithya Iyer, Lisa Kondrich, Nikhil~X. Bhattasali, and Lerrel Pinto.
\newblock Ruka: Rethinking the design of humanoid hands with learning.
\newblock In \emph{Robotics Science and Systems (RSS)}, 2025.

\end{thebibliography}
\clearpage
\appendix
\subsection{Tactile Feedback Module Details}
\label{app:fabrication}
\subsubsection{\textbf{Feedback Module Fabrication}}
The general fabrication workflow aligns with the methods established in \cite{FlatPanel}. However, a key distinction in our approach is the elimination of laser cutting. As discussed in \cref{subsec:tactile_design}, our design optimizations—specifically the increased inter-node spacing—allow the entire assembly to be handcrafted without relying on high-precision industrial machining.

\begin{figure}[!ht]
  \centering
  \includegraphics[width=0.48\textwidth]{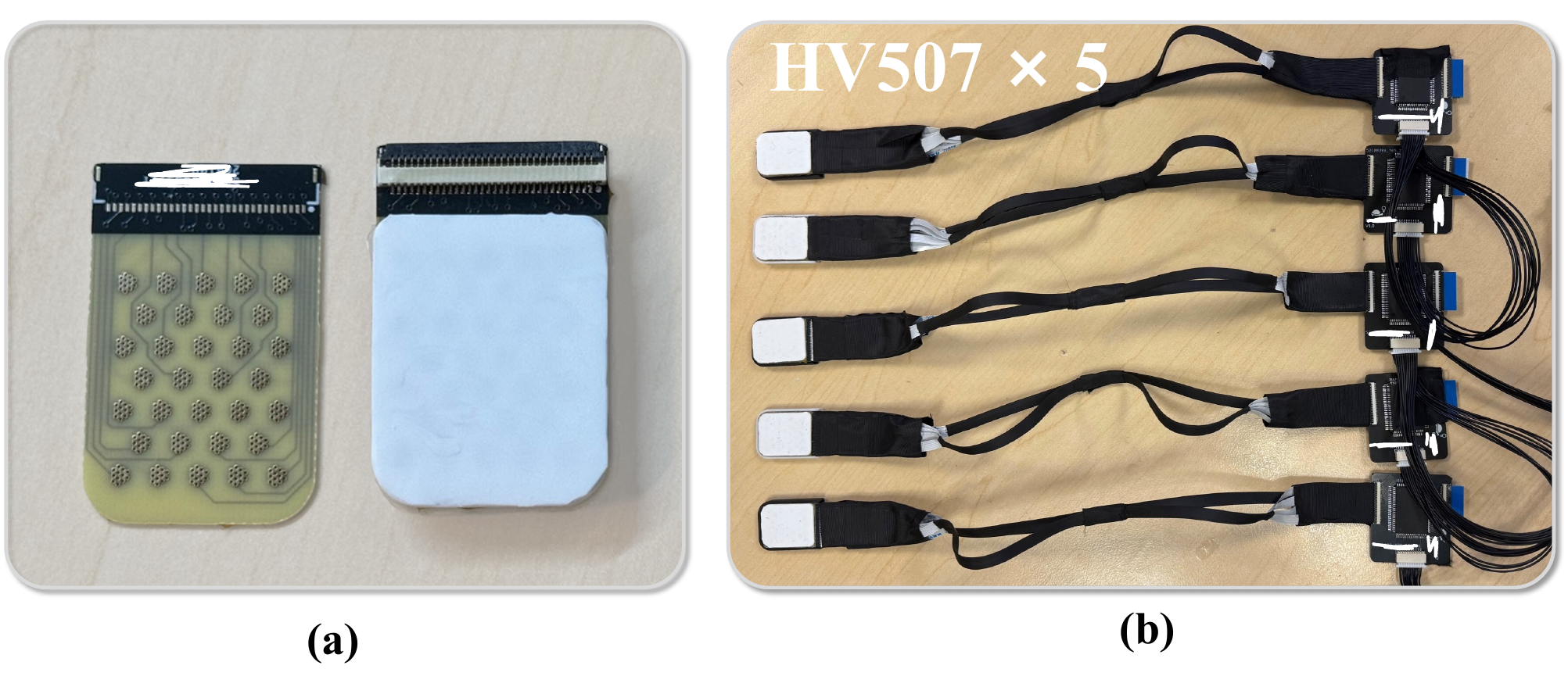}
  \vspace{-0.1in}
    \caption{\textbf{Tactile Feedback Module and Control Circuit.} (a) Top view of the Tactile Feedback Module and its single-layer PCB; (b) Five modules correspond to the five fingers, each independently driven by an HV507 chip.}
  \label{fig:app_eop}
    \vspace{-0.1in}
\end{figure}

The driving layer consists of two PCB layers bonded together using Pressure Sensitive Adhesive (PSA, 3M 468MP) and a PET spacer. Top-down views of the PCB layout and the assembled feedback module are illustrated in \cref{fig:app_eop}(a). The fabrication process is as follows:

\begin{itemize}[leftmargin=15pt, topsep=2pt, itemsep=1pt]
    \item \textbf{Spacer Preparation:} PSA sheets are applied to both sides of a PET film. Using a custom 3D-printed mold and a circular punch, through-holes are cut into the PET film to match the feedback via patterns on the PCBs, as shown in the PET layer of \cref{fig:design}(b).
    
    \item \textbf{Membrane Integration:} The prepared PET spacer is bonded to the bottom PCB layer. Glass fiber (GF/F) membranes ($1.6\,\text{mm}$ diameter), which are similarly cut using a circular punch, are then carefully placed into the 32 designated voids within the spacer.
    
    \item \textbf{Sealing:} The top PCB layer is aligned and bonded onto the assembly. Constant pressure is applied during curing to ensure a hermetic seal and effective isolation between adjacent fluidic channels.
\end{itemize}

To integrate the soft structure with the rigid driving layer, we use Sil-Poxy adhesive to bond the Ecoflex™ 00-30 output and reservoir components to the PCB. Once the module is fully bonded and cured, Propylene Carbonate (PC) is introduced as the dielectric working fluid. Using a syringe, PC is injected into the reservoir; simultaneously, trapped air is evacuated from the chamber to ensure a bubble-free hydraulic system.

\subsubsection{\textbf{Control Circuitry}}
Each feedback module features a 32-taxel tactile array requiring independent voltage regulation for both the top and bottom electrodes, totaling 64 control channels. This configuration enables tri-state control for each taxel: active protrusion (+), neutral (0), and retraction (-). As illustrated in \cref{fig:app_eop}(b), each module is driven by a single HV507 (Microchip Technology), a 64-channel high-voltage serial-to-parallel converter with push-pull outputs. The HV507 supports a daisy-chain configuration and features a Latch Enable (LE) function, allowing the states of multiple HV507 chips to be synchronized and controlled using a minimal number of digital I/O pins.

\subsection{Retargeting Details}

\subsubsection{\textbf{Hand Retargeting}}
The tactile glove provides its joint angle information $\theta_g$ at a frequency of $260$~Hz. 
Through forward kinematics, we can obtain the poses $(p_g, q_g)$ of the glove's end-links relative to the wrist frame.
To match the size of the dexterous hand, we scale $p_g$ by a factor $\lambda_f$, defined as the ratio between the dexterous hand's finger length and the operator's corresponding finger length, yielding the target fingertip positions $p_{gt}$, while keeping the target orientations unchanged ($q_{gt}=q_g$).

\begin{figure}[!ht]
  \centering
  \includegraphics[width=0.48\textwidth]{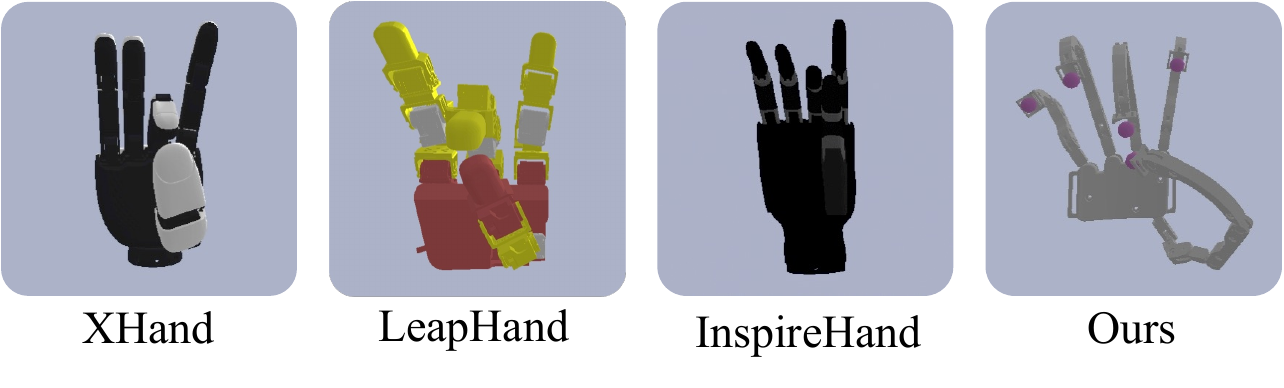}
  \vspace{-0.1in}
   \caption{\textbf{Hand retargeting performance.} The optimization-based method is applied to three distinct robotic hands: XHand, LeapHand and InspireHand. Despite differences in kinematics and dimensions, all hands successfully reproduce the target gesture captured from \ours.}
  \label{fig:hand_retarget}
    \vspace{-0.1in}
\end{figure}

Preserving thumb--finger opposition is crucial for successful manipulation during retargeting.
Therefore, we incorporate an opposition constraint to encourage the dexterous hand to reproduce the operator's finger-closing behavior.
Specifically, we formulate the following optimization problem:
\begin{align}
\min_{\theta}\quad
& w_1 \sum_i \left\| p_i(\theta)-p_{gt,i} \right\|_2^2
+ w_2 \sum_i \left\| R_i(\theta)-R_{gt,i} \right\|_F^2 \notag \\
&\quad + \alpha \left\| \theta-\theta_{\mathrm{last}} \right\|_2^2
\label{eq:hand_retarget}\\
&\quad + \sum_j \mathrm{weight}_j
\Big( d_{\mathrm{thumb},j}(\theta)-d_{\mathrm{thumb},j}^{gt} \Big)^2 \notag \\
\text{s.t.}\quad
& \theta_k^{\min} \le \theta_k \le \theta_k^{\max},\quad \forall k . \notag
\end{align}
where $p_i(\theta)$ and $R_i(\theta)$ denote the position and orientation (as a rotation matrix) of the $i$-th fingertip link under joint angles $\theta$; $d_{\mathrm{thumb},j}(\theta)$ is the distance between the thumb and the $j$-th finger; 
and $\mathrm{weight}_j$ is a piecewise function of the current thumb--finger distance $d_{\mathrm{thumb},j}$: it equals $0$ when $d_{\mathrm{thumb},j}$ exceeds an upper threshold $d_{\max}$, increases from $0$ to $w_{\max}$ as $d_{\mathrm{thumb},j}$ decreases from $d_{\max}$ to a lower threshold $d_{\min}$, and remains clamped at $w_{\max}$ when $d_{\mathrm{thumb},j} \le d_{\min}$.
The bounds $\theta_k^{\min}$ and $\theta_k^{\max}$ denote the joint-angle limits for each component of $\theta$.
Solving \cref{eq:hand_retarget} yields the dexterous-hand joint command $\theta_t$ at time step $t$, which is then sent to the hand controller to track the retargeted motion. We can solve this optimization problem at approximately 3000~Hz on an i7-12700 CPU, which is sufficient for real-time control. The qualitative results of the proposed retargeting framework are illustrated in \cref{fig:hand_retarget}.

\subsubsection{\textbf{Arm Retargeting}}
We implement arm retargeting by mapping the operator's wrist pose directly to the robot's end-effector. As detailed in \cref{sec:arm_retargt}, we employ two distinct interfaces for pose acquisition: the HOMIE exoskeleton and the HTC VIVE trackers. Below, we describe the implementation details for each setup.
\mbox{}\par \textbf{HOMIE Exoskeleton:} We adopt the exoskeleton interface design from HOMIE~\cite{ben2025homie} and integrate it into our \ours system. By utilizing Forward Kinematics (FK), we acquire both the arm joint poses and the wrist pose.

For the Unitree G1, our setup remains consistent with the original HOMIE implementation. We employ a 7-DoF isomorphic exoskeleton for the right arm, which is directly interfaced with \ours. We do not extend the FK-based exoskeleton adaptation of \ours to the UR5e. For exoskeleton-based teleoperation of the UR5e, we refer readers to the implementation in GELLO~\cite{wu2024gello}.
\mbox{}\par \textbf{HTC VIVE Trackers:} To acquire precise wrist pose information, we utilize the Vive tracking system, which consists of two base stations (infrared (IR) emitters) and a tracker. The system employs infrared laser scanning to achieve high-precision localization of the tracker. 
Each tracker provides 6-DoF pose data relative to the base station coordinate system, including 3D position \( p \in \mathbb{R}^3 \) and orientation represented as a quaternion \( q \), at a sampling frequency of 250 Hz. 

In our setup, we attach a Vive tracker to the operator's glove to capture accurate wrist pose.
We first transform the pose information into a coordinate system aligned with the robot's base frame and record the initial pose \( p_0, q_0 \). At each time step \( t \), we compute the incremental change in pose relative to the initial pose as \( \Delta p_t, \Delta q_t \).
Since the operator's arm movement range typically exceeds that of the robot, we introduce a scaling factor \( \lambda \) to adjust the positional increment \( \Delta p_t \).

For the Unitree G1, we apply this scaled positional increment and the rotational change to the initial pose of the robot's end-effector \( p_{r0}, q_{r0} \) to obtain the target pose \( p_{rt}, q_{rt} \).
Finally, we map the target pose to the robot's joint space using inverse kinematics. Specifically, we formulate an optimization problem:
\begin{align}
\min_{\theta}\quad
& w_1 \| p(\theta) - p_{rt} \|^2
+ w_2 \| R(\theta) - R_{rt} \|_F \notag \\
&\quad + \alpha \| \theta - \theta_{last} \|^2
\label{eq:arm_retarget_ik}\\
\text{s.t.}\quad
& \theta_k^{\min} \le \theta_k \le \theta_k^{\max},\quad \forall k . \notag
\end{align}
where \( p(\theta) \) denotes the end-effector position of the robot at joint angles \( \theta \), and \( \alpha \) is the weight of the regularization term. The bounds $\theta_k^{\min}$ and $\theta_k^{\max}$ denote the joint-angle limits for each component of $\theta$.
By solving this optimization problem, we obtain the robot's joint angles \( \theta_t \) at the current time step \( t \), enabling pose tracking of the robot's end-effector through the robot's control interface.

For the UR5e, we similarly add the increments \( \Delta p_t, \Delta q_t \) to the initial pose of the arm's end-effector \( p_{r0}, q_{r0} \) to obtain the target pose \( p_{rt}, q_{rt} \), and then utilize the UR5e's built-in interface for target pose tracking.

\subsubsection{\textbf{Key Note}} A critical implementation detail concerns signal interference. The RealSense D455 camera utilizes an active IR projector for depth estimation. When its projection field overlaps with the VIVE base stations, the IR emission causes crosstalk, resulting in tracking jitter or severe localization errors. To avoid this, we strategically adjust the spatial positioning of the base stations to eliminate signal overlap, successfully ensuring robust tracking performance.

\subsection{Experimental Details}
\subsubsection{User Study}
We use the \textbf{Contact Shape Discrimination} task to illustrate the protocol. The experimental workflow follows the procedures described in \cref{sec:Contact_shape}. Participants are blindfolded and wear noise-canceling headphones to eliminate visual and auditory cues. Since they cannot actively align the robotic finger with the target shapes, a second experimenter positions the geometries beneath the Inspire Hand’s index finger.

In each trial, the second experimenter shuffles and randomly selects one geometry, pressing it against the sensor with sufficient force to generate clear tactile features. The second experimenter also monitors signal quality in real-time to ensure proper data capture, maintaining contact for 5 seconds. Upon removal, the participant immediately reports the perceived shape relying solely on tactile feedback. This process continues until all four geometries are tested, completing one experimental block. Each participant performs 5 blocks (20 trials total). With 5 participants, a total of 100 trials are conducted. The results are presented in \cref{fig:user_results}.

\begin{figure}[!ht]
  \centering
  \includegraphics[width=0.48\textwidth]{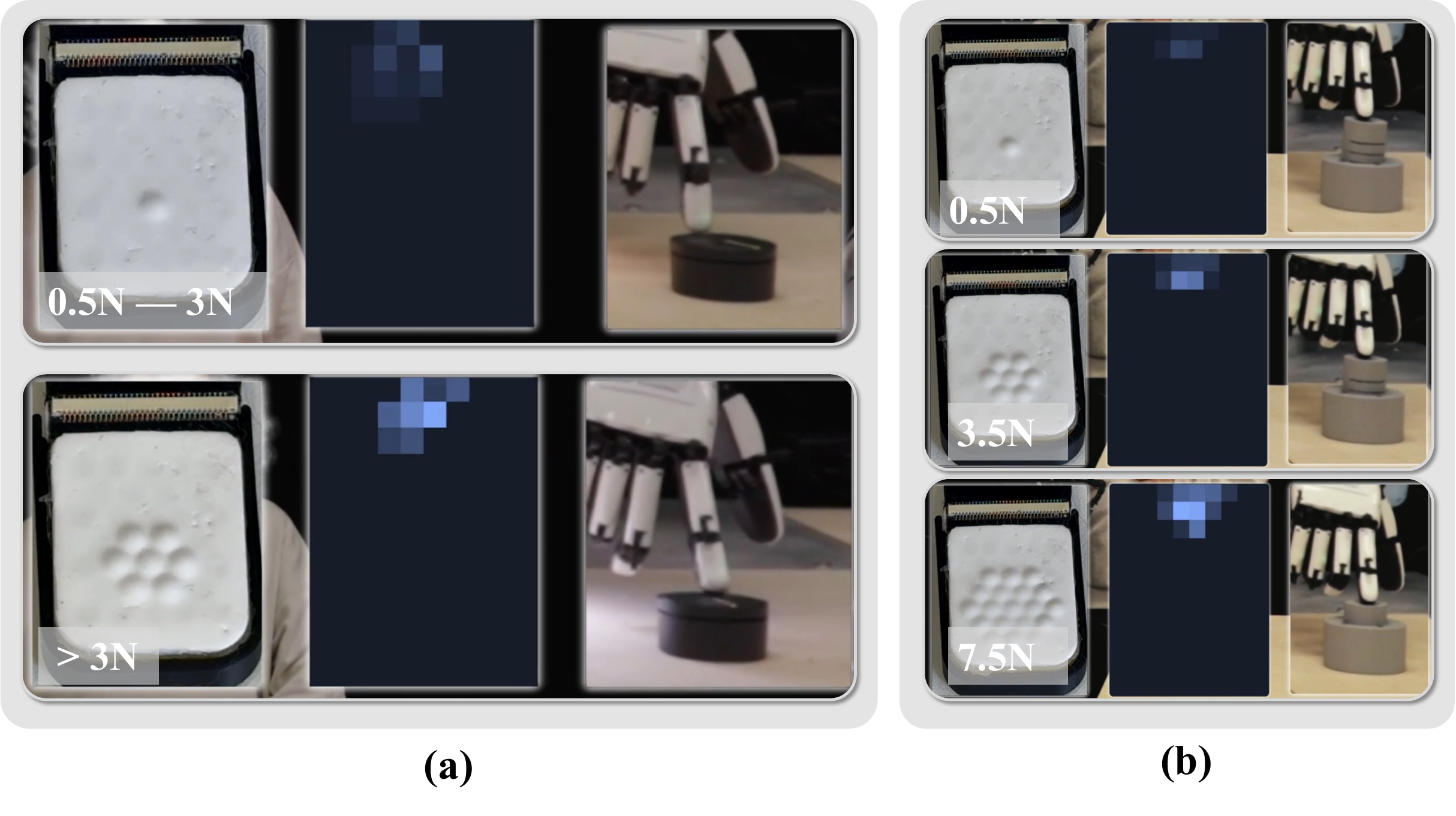}
  \vspace{-0.1in}
    \caption{\textbf{Illustration of teleoperation tasks in Press Mode.} (a) Multimodal Control of a Smart Desk Lamp. (b) Multi-level Spring Compression.}
  \label{fig:app_exp1}
    \vspace{-0.1in}
\end{figure}

\subsubsection{\textbf{Teleoperation Experiments Details}}
The general experimental setup follows the description in \cref{sec:teleop_perfo}. For comparative analysis, we select the HOMIE system~\cite{ben2025homie} as the baseline—specifically, the HOMIE Glove integrated with a VIVE Tracker. This baseline device is a 15-DoF exoskeleton glove that employs single-axis Hall sensors for joint encoding but lacks tactile feedback capabilities.

\begin{figure}[!ht]
  \centering
  \includegraphics[width=0.43\textwidth]{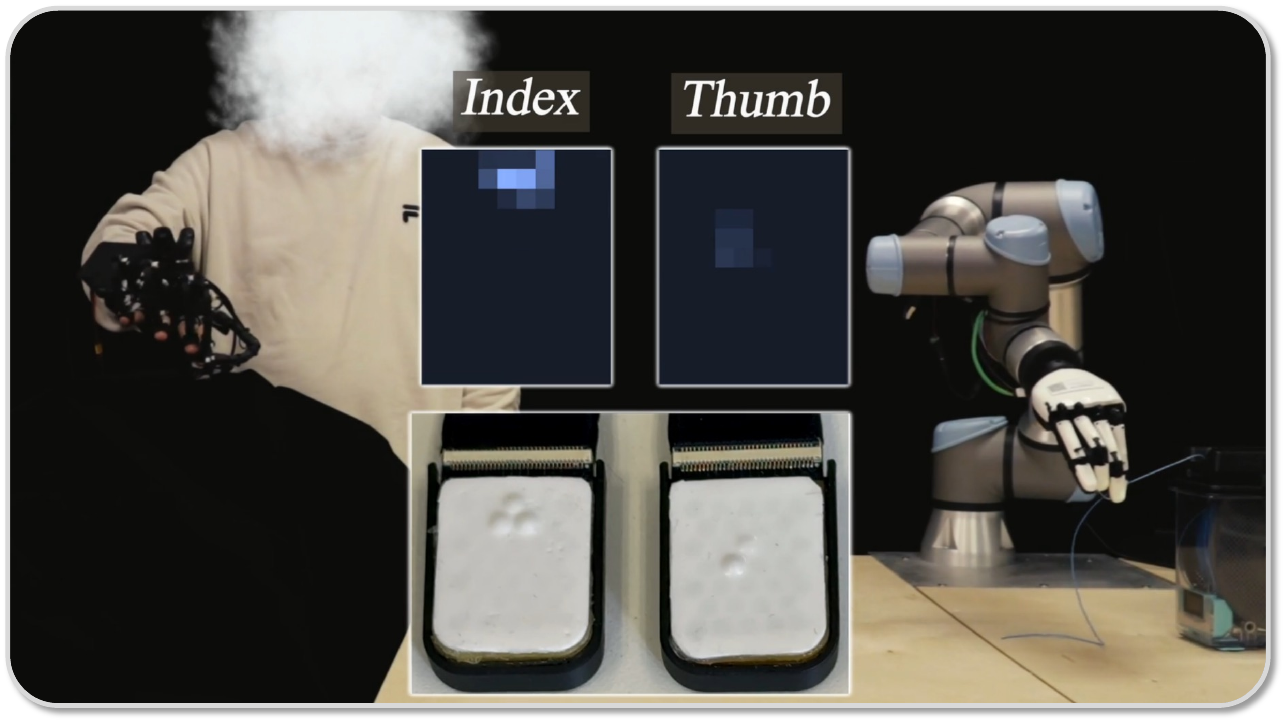}
   \caption{\textbf{Illustration of Filament Pinching and Pull-out Task.} The schematic shows the thumb and index finger pinching the filament.}
  \label{fig:app_exp2}
    \vspace{-0.1in}
\end{figure}

\begin{itemize}[leftmargin=15pt, topsep=2pt, itemsep=1pt]
    \item \textbf{Multimodal Control of a Smart Desk Lamp} – Supplementary to the protocol, Stages 2 and 3 require the operator to perform multiple rotational clutching motions (typically 3--5 repetitions) to traverse the full control range. The trial ends immediately upon reaching the target state, defined as the warmest CCT or minimum brightness. A trial is classified as a failure if any incorrect regulation occurs, particularly in Stage 3: if the downward force exceeds 3N during rotation, the system unintentionally switches to the CCT mode of Stage 2, resulting in an immediate failure. Since \ours is configured in \textbf{Press Mode}, the tactile protrusion is modulated based on real-time pressure; specifically, \ours protrudes by 1 point (0.5N-3N) and by 7 points ($>3$N), as illustrated in \cref{fig:app_exp1}(a).
    \item \textbf{Filament Pinching and Pull-out} – This task entails two consecutive pinch-and-pull actions. The trial fails if either attempt is unsuccessful. Failure modes include missing the grasp initially (grasping air) or the filament slipping from the fingertips during pull-out. The schematic of the thumb and index finger pinching the filament is shown in \cref{fig:app_exp2}.
    
    \item \textbf{Multi-level Spring Compression} – Whether blindfolded or sighted, the operator must verbally report the perceived force level upon engagement and maintain pressure for 2 seconds. A trial fails if the actual compression depth mismatches the reported level—specifically, if the visual markers on the spring-loaded button fall short of or exceed the target lines. The three pressure levels and corresponding feedback are detailed in \cref{fig:app_exp1}(b).
\end{itemize}

\subsection{LeapTac Design}
The LeapHand is extensively employed in robotic manipulation research; however, it notably lacks integrated tactile sensing capabilities. To address this limitation, as detailed in \cref{subsec:tac-eval}, we present \textit{LeapTac}, a custom-designed resistive tactile array tailored for the LeapHand fingertip. The fundamental sensing principle of LeapTac aligns with the piezoresistive approach described in \cite{Huang20243DViTacLF}.

\begin{figure}[!ht]
  \centering
  \includegraphics[width=0.48\textwidth]{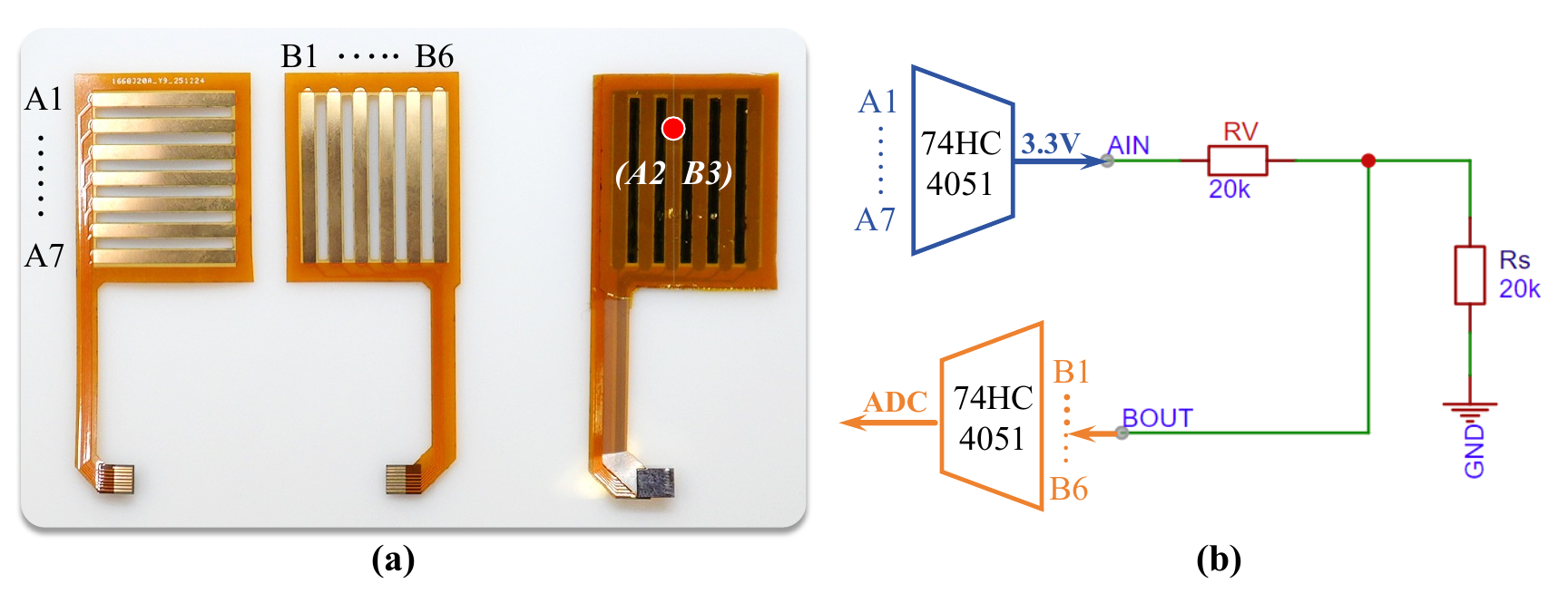}
  \vspace{-0.1in}
   \caption{\textbf{Physical prototype and schematic of the LeapTac sensor.} (a) The two-layer Flexible Printed Circuit (FPC) electrodes and the fully assembled LeapTac module. (b) The circuit schematic illustrating the readout mechanism for individual sensing points (taxels).}
  \label{fig:leaptac}
    \vspace{-0.1in}
\end{figure}

Structurally, the sensor comprises a piezoresistive Velostat sheet sandwiched between two flexible printed circuit (FPC) layers containing orthogonal conductive traces. The top layer features transverse traces while the bottom layer features longitudinal traces, forming a grid architecture where each intersection point functions as an individual sensing element, or \textit{taxel}, as illustrated in \cref{fig:leaptac}(a). This sandwich structure ensures that pressure applied to the surface compresses the Velostat locally, reducing its electrical resistance at the contact point. To seamlessly accommodate the curved geometry of the LeapHand fingertip, the FPC electrodes are designed with a trace width of 2 mm and a center-to-center spacing of 4 mm. This configuration yields a spatial resolution of $7 \times 6$ taxels per fingertip, providing sufficient density for object localization and force estimation.

The electronic interface is designed to minimize wiring complexity while maximizing sampling speed. Data acquisition is managed by two 8-channel analog multiplexers (74HC4051) operating in a row-column scanning mode. As shown in the schematic in \cref{fig:leaptac}(b), one multiplexer controls the voltage driving lines (rows), while the other handles the signal readout lines (columns). To query a specific taxel ($R_i, C_j$), the driving multiplexer selectively applies a reference voltage $V_{cc}$ to row $R_i$. Current flows through the resistive film at the intersection to column $C_j$, which is connected via the readout multiplexer to a microcontroller's ADC through a voltage divider circuit. By measuring the output voltage, we can calculate the instantaneous resistance of the Velostat using Ohm's law. This resistance change is inversely proportional to the normal force applied at the specific coordinate ($R_i, C_j$), allowing the system to reconstruct a real-time pressure map of the contact area.

\end{document}